\documentclass[sn-mathphys-num]{sn-jnl}

\usepackage{multirow}%
\usepackage{amsmath,amssymb,amsfonts}%
\usepackage{amsthm}%
\DeclareMathAlphabet{\mathscr}{U}{rsfs}{m}{n}
\usepackage[title]{appendix}%
\usepackage{xcolor}%
\usepackage{textcomp}%
\usepackage{manyfoot}%
\usepackage{booktabs}%
\usepackage{algorithm}%
\usepackage{algorithmicx}%
\usepackage{algpseudocode}%
\usepackage{listings}%
\usepackage{newtxtext, newtxmath}
\usepackage{hyperref}
\usepackage{tabularx}
\usepackage{graphicx} 
\usepackage{color}                        
\usepackage[normalem]{ulem}  
\usepackage{geometry}        
\usepackage{xspace}  
\usepackage{threeparttable}    
\usepackage{subcaption} 
\usepackage[T1]{fontenc}

\hypersetup{colorlinks=true, linkcolor=blue, citecolor=blue, urlcolor=blue, bookmarksdepth=3}

\title[Article Title]{TabNSA: Native Sparse Attention for Efficient Tabular Data Learning}

\author[1]{\fnm{Ali} \sur{Eslamian}}
\author*[1,2]{\fnm{Qiang} \sur{Cheng}}

\affil[1]{\orgdiv{Department of Computer Science}, \orgname{University of Kentucky}, \orgaddress{\street{329 Rose Street}, \city{Lexington}, \postcode{40506}, \state{Kentucky}, \country{USA}}}
\affil[2]{\orgdiv{Institute for Biomedical Informatics}, 
\orgname{University of Kentucky}, \orgaddress{\street{800 Rose Street}, \city{Lexington}, \postcode{40506}, \state{Kentucky}, \country{USA}}}

\abstract{
Tabular data poses unique challenges for deep learning due to its heterogeneous feature types, lack of spatial structure, and often limited sample sizes. We propose TabNSA, a novel deep learning framework that integrates Native Sparse Attention (NSA) with a TabMixer backbone to efficiently model tabular data. TabNSA tackles computational and representational challenges by dynamically focusing on relevant feature subsets per instance. The NSA module employs a hierarchical sparse attention mechanism, including token compression, selective preservation, and localized sliding windows, to significantly reduce the quadratic complexity of standard attention operations while addressing feature heterogeneity. Complementing this, the TabMixer backbone captures complex, non-linear dependencies through parallel multilayer perceptron (MLP) branches with independent parameters. These modules are synergistically combined via element-wise summation and mean pooling, enabling TabNSA to model both global context and fine-grained interactions. Extensive experiments across supervised and transfer learning settings show that TabNSA consistently outperforms state-of-the-art deep learning models. Furthermore, by augmenting TabNSA with a fine-tuned large language model (LLM), we enable it to effectively address Few-Shot Learning challenges through language-guided generalization on diverse tabular benchmarks.

}
\begin{document}
\maketitle

\newpage

\section{Introduction}

Tabular data, prevalent in domains like finance, healthcare, and e-commerce, poses unique challenges due to its heterogeneity, lack of spatial or temporal structure, non-uniform feature interactions, and often limited sample sizes \cite{borisov2022deep}. Traditional models like Gradient Boosted Decision Trees (GBDT) \cite{lu2025large, chen2024dofen} excel in these settings, but deep learning models often struggle to match their performance.

In response to these challenges, researchers have proposed a variety of deep learning models tailored to tabular data. Some approaches adapt multilayer perceptrons (MLPs) with enhanced regularization or embedding mechanisms, while others introduce architectural innovations such as attentive feature selection, gating mechanisms, or sparsity-inducing modules. Notable examples include TabNet \cite{tabnet}, which utilizes attentive feature masks to model sparse feature dependencies, and DSelect-k \cite{hazimeh2021dselect}, which provides a differentiable relaxation of feature subset selection. More recently, Transformer-inspired models have been explored for capturing pairwise and higher-order feature interactions \cite{gorishniy2021revisiting}. While these methods have advanced the field, many remain computationally expensive, lack dynamic sparsity, or fail to generalize across tasks with varying data regimes, such as low-data or feature-incremental settings.

To bridge this gap, recent research has proposed deep learning architectures specifically tailored for tabular data. One such model, TabMixer \cite{eslamian2025tabmixer}, extends the MLP-Mixer architecture by employing channel-wise and token-wise mixing to capture global feature and instance dependencies. It achieves strong performance on supervised, transfer and incremental learning tasks while maintaining low computational complexity and robustness to missing values. However, a key limitation of TabMixer is that it processes all features uniformly, without distinguishing instance-specific feature importance, which potentially leads to redundancy and reduced interpretability.

In parallel, Native Sparse Attention (NSA) \cite{yuan2025native} has demonstrated the effectiveness of sparse attention mechanisms in reducing computational overhead while preserving performance in sequence modeling. Inspired by this, we propose TabNSA, a new deep learning framework that integrates NSA with the TabMixer backbone to address both computational and modeling challenges in tabular learning.

TabNSA introduces a hierarchical sparse attention mechanism that dynamically selects a relevant subset of features for each instance, effectively reducing redundancy and highlighting instance-specific dependencies. This coarse-to-fine selection is followed by TabMixer’s MLP-based token and channel mixing, which efficiently models the complex interactions of the selected features. NSA enhances the model adaptability and scalability, while TabMixer provides expressive capacity to capture non-linear dependencies. Their integration results in a lightweight yet powerful architecture capable of handling high-dimensional, heterogeneous tabular data.

Adapting NSA to tabular data involves treating features as tokens, shifting the attention paradigm from sequences (e.g., time steps or words) to feature dimensions. This design choice is motivated by the observation that in tabular datasets, not all features are equally informative for every instance, and their interactions are often sparse and heterogeneous. Treating features as tokens allows the model to dynamically allocate attention to a subset of relevant dimensions, enabling instance-specific modeling of feature relevance. The hierarchical sparse attention of the NSA, comprising token compression, selection, and localized attention, naturally aligns with these properties. In particular, the token compression mechanism mitigates redundancy among correlated features, while selective preservation allows the model to focus computational resources on informative features, addressing both heterogeneity and sample efficiency. Compared to standard dense attention, which treats all features uniformly and scales quadratically with feature count, NSA provides a principled and scalable alternative that is better suited for high-dimensional, noisy tabular inputs. In our ablation studies (see Experiment section), we empirically confirm that NSA outperforms full-attention variants in both accuracy and efficiency across diverse tabular benchmarks.


In summary, this paper makes the following contributions:

\begin{itemize}
\item \textbf{{TabNSA Architecture:}} We propose TabNSA, a novel deep learning model integrating NSA with a TabMixer-based representation to dynamically select and process instance-specific feature subsets. This fusion effectively captures both global and local feature interactions, enhancing efficiency and expressiveness for tabular data.

\item \textbf{{Efficient Hierarchical Sparse Attention:}} TabNSA employs effective hierarchical sparse attention mechanisms, significantly reducing attention complexity without sacrificing performance. This enables the model to scale efficiently to large tabular datasets.

\item {\textbf{Efficiency and Generalization:}} By leveraging sparse attention and hierarchical feature processing, TabNSA efficiently captures long-range dependencies and feature-wise interactions. This contributes to strong generalization across various tasks and facilitates insightful feature importance analysis, enhancing interpretability.

\item \textbf{{Enhanced Few-Shot and Transfer Learning Capabilities:}} Through integration with large language models such as Gemma, TabNSA demonstrates superior performance in few-shot and low-resource learning scenarios, significantly outperforming existing baselines.

\item \textbf{{State-of-the-Art Performance:}} Extensive empirical evaluation across supervised, transfer, and few-shot learning tasks demonstrates that TabNSA consistently achieves state-of-the-art performance, surpassing leading deep learning architectures, particularly in low-resource and transfer settings.

\end{itemize}

To situate TabNSA within the broader literature, we next review related work in tabular learning and sparse attention mechanisms.

\section{Related Work}

Existing tabular learning methods aim to address various challenges associated with diverse datasets. These include situations where the feature sets differ between training and test data, limited or missing labels, and the potential for new features to emerge over time \cite{maqbool2024model}. We will briefly review related work in the following. 

{\textbf{Classical techniques}} include parametric methods like Logistic Regression and non-parametric approaches such as K-Nearest Neighbors, Gradient Boosting, and Decision Trees \cite{Moderndeeplearning}. Various models have been developed, including logistic regression, XGBoost \cite{chen2016xgboost, zhang2020customer}, and Multi-Layer Perceptrons. 

{\textbf{Deep learning-based approaches}} have emerged to address the growing complexity and high-dimensionality of modern tabular datasets. These methods can automatically learn hierarchical representations and capture intricate patterns within the data \cite{Deeplearning}. Deep cross networks (DCN) \cite{wang2017deep} combine a deep network that learns high-order feature interactions with a cross-network that automatically applies feature crossing, efficiently capturing bounded-degree feature interactions. TabMixer \cite{eslamian2025tabmixer} uses an enhanced MLP-Mixer architecture. Its design effectively extracts attention-aware features and captures complex interactions across tabular features and examples, even extending to dependencies between different tables.

{\textbf{Attention-based models}}, in particular, have gained popularity as they dynamically focus on the most relevant parts of the input data, although this can increase computational requirements.
TabNet \cite{tabnet} is a attention-based mechanisms for tabular data processing. It concentrates on the most significant features at each decision stage, enabling efficient learning and providing interpretable feature attributions. Several models have been developed using Transformers as their foundational components. AutoInt \cite{song2019autoint} employs Transformers to determine the significance of diverse input features. TransTab \cite{transtab} utilizes Transformers to handle tasks such as learning from multiple tables with partially overlapping columns. Similarly, TabTransformer \cite{tabtransformer} employs self-attention-based Transformers to enhance feature representations, surpassing classical methods in terms of accuracy, exhibiting robustness to missing data, and introducing a novel pre-training approach for semi-supervised learning. SAINT \cite{SAINT} uses a hybrid attention mechanism, focusing on both rows and columns with enhanced embeddings. It introduces a novel intersample attention approach and contrastive pre-training, outperforming gradient boosting methods.

\textbf{Sparse Attention Mechanisms.} Sparse attention has emerged as a powerful paradigm for improving the efficiency and interpretability of deep models, particularly in high-dimensional settings. In NLP, methods such as Longformer, BigBird, and Routing Transformer \cite{beltagy2020longformer, zaheer2020bigbird, roy2021efficient} introduce sparsity into the attention matrix to handle long sequences more efficiently without sacrificing contextual understanding. In computer vision, sparse attention has been used to focus on salient spatial regions while reducing computation, as seen in models like SparseViT and TokenLearner \cite{mano2021tokenlearner}. The recent Natively Sparse Attention (NSA) framework \cite{yuan2025native} extends this idea by learning instance-specific sparse attention patterns through token compression, selection, and local context modeling, achieving competitive performance with reduced cost in language and vision tasks. Unlike fixed sparse patterns (e.g., in BigBird), NSA learns data-driven sparsity patterns per instance, which is critical in tabular data where relevant features vary

We draw inspiration from these advances and adapt NSA to tabular data, where high dimensionality, feature redundancy, and heterogeneous feature relevance across instances present similar challenges. By treating features as tokens, sparse attention enables TabNSA to dynamically select a subset of informative features per instance—mirroring the instance-adaptive attention behavior that proved successful in sequence and image domains. This adaptation helps improve computational scalability and enhances interpretability for tabular learning tasks.

\section{Background}

To address the computational inefficiency of standard attention mechanisms in long-context modeling for tabular data processing, we adapt Native Sparse Attention, a natively trainable sparse attention architecture~\cite{NSA-1, NSA-2}. Additionally, we incorporate TabMixer~\cite{eslamian2025tabmixer}, which captures complex patterns and global dependencies inherent in tabular data, where feature relationships are often non-linear and interdependent. 

\subsection{Native Sparse Attention}
For tabular data learning, NSA can be adapted by treating each feature as a token and learning the relationships between these features through its sparse attention mechanisms.
In the architecture, each sample’s feature vector is first embedded and then projected into queries, keys, and values:
\begin{equation}
q = W_q X, \quad k = W_k X, \quad v = W_v X.
\end{equation}
This corresponds to the standard attention formulation. Here, $W_q$, $W_k$ and $W_v$ are learned projection matrices. Given an input sequence with query $q_t$, keys $K_t$, and values $v_t$ for $t$-features the full attention computes the output as:
\begin{equation}
\begin{aligned}
    o_t &= \text{Attn}(q_t, k_{1:t}, v_{1:t}) = \sum_{i=1}^{t} \alpha_{t,i} v_i, \\
    \text{with} \quad \alpha_{t,i} &= 
    \frac{\exp \left( \frac{q_t k_i}{\sqrt{d_k}} \right)}
    {\sum_{j=1}^{t} \exp \left( \frac{q_t k_j}{\sqrt{d_k}} \right)}.
\end{aligned}
\end{equation}

NSA replaces the original key value pairs with a more compact representation $\tilde{k}_t \text{ and } \tilde{v}_t$ constructed as:
\begin{equation}
\tilde{k}_t = f_K(q_t, k_{1:t}, v_{1:t}), \quad
\tilde{v}_t = f_V(q_t, k_{1:t}, v_{1:t})
\end{equation}

The functions $f_K$ and $f_V$ implement three distinct remapping strategies $\mathcal{C} = \{cmp, slc, win\}$ representing compression, selection, and sliding window, respectively:

\begin{enumerate}

\vspace{2mm}
\item  \textbf{Token Compression:} Aggregates sequential blocks of keys or values into block-level representations. The compressed key representation is defined as:
\begin{equation}
\tilde{k}_t^{\text{cmp}} = f_K^{\text{cmp}}(k_{1:t}) = \left\{ \varphi \left( k_{id+1:id+l} \right) \middle| 0 \leq i \leq \left\lfloor \frac{t - L}{d} \right\rfloor \right\}
\end{equation}
where $l$ is the block length, $d$ is the sliding stride between adjacent blocks, and $\varphi$ is a learnable MLP with intra-block position encoding.

\vspace{2mm}
\item \textbf{Token Selection:} 
Selectively preserves individual keys and values. The selection strategy processes key and value sequences in spatial continuous blocks. The importance scores for selection blocks are derived from the attention computation of compression tokens:
\begin{equation}
    \mathbf{p}_t^{\text{cmp}} = \text{Softmax}\left(\mathbf{q}_t^T \tilde{k}_t^{\text{cmp}}\right)
\end{equation}

$\mathbf{p}_t^{\text{cmp}}$ denote the attention scores between $q_t$ and the compressed keys. 

Let $l^\prime$ denote the selection block size. By using the same blocking pattern for both compression and selection blocks, we can directly obtain the importance scores for the selection blocks $\mathbf{p}_t^{\text{slc}}=\mathbf{p}_t^{\text{cmp}}$. When the blocking schemes vary, we derive the importance scores for selection blocks based on their spatial relationships.So, the selection score for block $j$ can be computed as:
\begin{equation}
\mathbf{p}_t^{\text{slc}}[j] = \sum_{m=0}^{\frac{l'}{d} -1} \sum_{n=0}^{\frac{l}{d} -1} 
\mathbf{p}_t^{\text{cmp}} \left[ \frac{l'}{d} j - m - n \right]
\end{equation}

where $\left[ \cdot \right]$ denotes the indexing operator for accessing vector element.

\begin{equation}
\mathcal{I}_t = \{ i \mid \operatorname{rank}(\mathbf{p}_t^{\text{slc}'}[i]) \leq n \}
\end{equation}

\begin{equation}
\tilde{k}_t^{\text{slc}} = \operatorname{Cat} \left[ \{ k_{i l' + 1 : (i+1) l'} \mid i \in \mathcal{I}_t \} \right],
\end{equation}

where rank$\left(\cdot\right)$ denotes the ranking position in descending order, with $rank=1$ corresponding to the highest score, $\mathcal{I}_t$ is the set of selected blocks indices, $\text{Cat}$ denotes the concatenation operation. Similarly, the fine-grained value can be expressed for $\tilde{v}_t^{\text{slc}}$.

\vspace{2mm}
\item \textbf{Sliding Window:} To handle local patterns effectively, a fixed-size sliding window of recent tokens is maintained:
\begin{equation}
\tilde{k}_t^{\text{win}} = k_{t-w:t}, \quad
\tilde{v}_t^{\text{win}} = v_{t-w:t}
\end{equation}

After the sparse attention module combines the outputs from the different branches, the resulting representation is processed by further linear layers (and activation functions like GELU) to produce the final output as follows:

\begin{equation}
Y = \sum_{c \in \mathcal{C}} g_t^c \cdot \text{Attn}(q_t, \tilde{K}_t^c, \tilde{V}_t^c)
\end{equation}

The gate score, $g^c_t \in [0,1]$, is obtained from the input features using sigmoid activation and an MLP.
\end{enumerate}

\subsection{TabMixer}
The TabMixer block \cite{eslamian2025tabmixer} is an enhanced version of the standard MLP-Mixer block, designed to process information through three main branches: a residual connection, a central pathway for channel-wise mixing, and a parallel path for token-wise mixing.

Unlike the standard MLP-Mixer which uses tied parameters in its MLPs, the TabMixer block utilizes parallel MLPs with independent learning parameters. This parallel structure, combined with an additional SiLU non-linear activation function, allows the model to more effectively capture complex, non-linear, and global dependencies within tabular data by exploring interactions across both features and examples simultaneously.

The block's operation can be described mathematically. For an input matrix $X \in \mathbb{R}^{B \times N}$ the output Z is computed as follows:

\begin{equation}
    Z = \text{SiLU} \left[\text{GeLU}\left( \text{MLP}_{1}(X^T) \right)^T \odot \text{MLP}_{2}(X) \right] + X
\end{equation}

\[
\text{MLP}_{1,2}(X) = W_{1,2} \text{LayerNorm}(X) + b_{1,2}
\]

The final output produces attention-aware features through additive adjustments. 

\subsection{Gemma}
Gemma \cite{Gemma} is a lightweight, open-source family of large language models (LLMs) developed by Google DeepMind, leveraging the technology behind Gemini models \cite{Gemma2}. Built upon a decoder-only Transformer architecture, Gemma models are designed for computational efficiency while maintaining strong performance across various natural language tasks. The Gemma-1.1-2B-IT variant, specifically employed in our study, features 2 billion parameters and has undergone instruction tuning. Instruction tuning enhances the model's capability to interpret natural language prompts, making it particularly effective for zero-shot or few-shot learning scenarios\cite{Gemma3} \cite{sanchez2024generative}. This capability is instrumental in our approach, where structured tabular data is transformed into textual prompts, harnessing Gemma’s advanced text representation and reasoning abilities for encoding tasks. Gemma's recent enhancements include extended context lengths, multilingual support, and a novel local-global attention mechanism for memory efficiency.

We selected the Gemma family of LLMs due to its strong performance on open-ended reasoning tasks and its flexibility in adapting to various downstream applications with minimal fine-tuning. Unlike encoder-only models such as BERT or RoBERTa, which are optimized for classification and masked language modeling, Gemma follows a decoder-style architecture more aligned with instruction tuning and generative few-shot prompting. This structure allows Gemma to better leverage natural language context and examples when guiding few-shot learning. In our preliminary tests, Gemma outperformed BERT and RoBERTa variants on few-shot tabular classification when using text-formatted input prompts, especially in extremely low-data regimes.

In our methodology, we freeze most encoder parameters of the pretrained Gemma to preserve generalized linguistic features and selectively fine-tune only the last Transformer layers. The resultant embeddings undergo a dual-pooling and linear-adaptation process before integration into our TabNSA classification module. This combined approach leverages Gemma’s pretrained linguistic power to ensure robust generalization and efficiency in few-shot learning contexts.
In the next section,  we describe how NSA’s sparse mechanisms address feature redundancy and TabMixer’s nonlinear mixing captures complex interactions, forming TabNSA.

\section{Method}
\subsection{Adapting NSA to Tablular Data}
First, we have adapted vanilla NSA to tabular data. The adaptation is detailed in Table~\ref{tab:nsa_adaptation}, which illustrates the conceptual mapping between the original use of NSA in sequence modeling and its adaptation for feature-wise modeling in tabular data. 

\begin{table}[htbp]
\centering
\resizebox{\linewidth}{!}{
\begin{tabular}{|p{4cm}|p{5cm}|p{5cm}|}
\hline
\textbf{Aspect} & \textbf{NSA in Sequence Modeling} & \textbf{NSA in Tabular Data (TabNSA)} \\
\hline
\textbf{Token definition} & Each token represents a time step or word in a sequence. & Each token corresponds to a feature (i.e., a column in the table). \\
\hline
\textbf{Input structure} & Ordered sequence of elements (e.g., text or time series). & Unordered, heterogeneous set of features. \\
\hline
\textbf{Attention focus} & Focuses on relevant past or future tokens. & Focuses on the most informative features per instance. \\
\hline
\textbf{Token compression} & Reduces temporal redundancy (e.g., repeated words or uninformative frames). & Reduces feature redundancy (e.g., correlated or irrelevant features). \\
\hline
\textbf{Local sliding window} & Preserves local temporal patterns. & Captures local dependencies among related feature groups. \\
\hline
\textbf{Sparse selection effect} & Enables efficient modeling of long sequences. & Enables efficient learning on high-dimensional, sparse tabular inputs. \\
\hline
\textbf{Interpretability} & Highlights influential time points or tokens. & Highlights per-instance key features, aiding interpretation. \\
\hline
\end{tabular}}
\caption{Conceptual adaptation of NSA from sequence modeling to tabular data modeling.}
\label{tab:nsa_adaptation}
\end{table}

\subsection{TabNSA Model Description} \label{Model_description}
Next, to effectively capture both long-range dependencies and feature-wise interactions in tabular data, we introduce \textbf{TabNSA}, a hybrid neural architecture that integrates \textit{Native Sparse Attention} (NSA) with \textit{TabMixer}. The model is designed to efficiently process tabular inputs by leveraging sparse self-attention for context-aware representation and TabMixer for non-linear, cross-feature interaction modeling.

\begin{figure*}[!t]
\centering
\includegraphics[width=\linewidth]{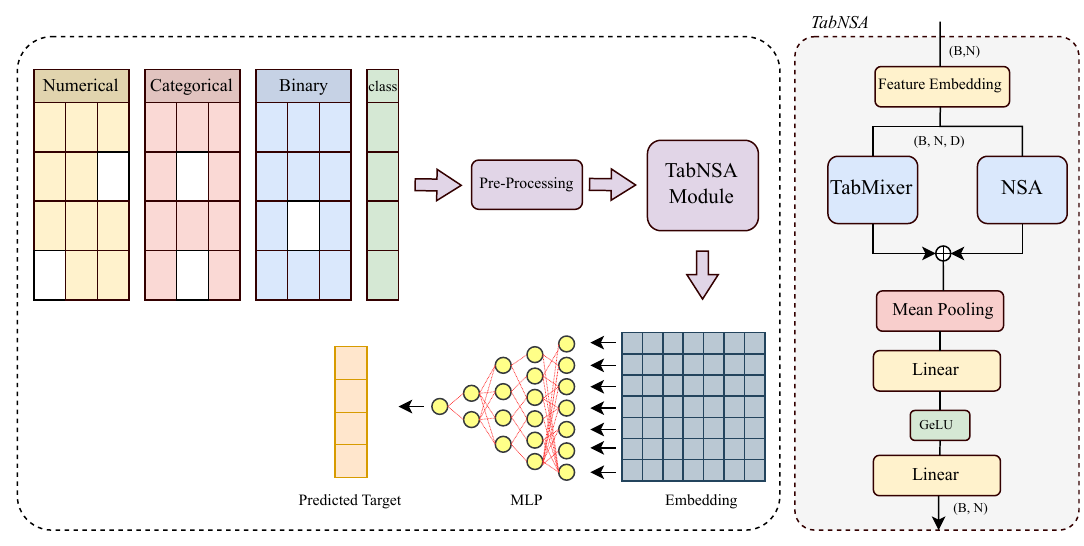}
\caption{Overview of the TabNSA model architecture. The input is a tabular feature vector of shape $(\textit{batch\_size}, \textit{NUM\_TOKENS})$. Each scalar feature is projected into a high-dimensional embedding space via a linear layer. The resulting token-feature matrix is processed in parallel by a Sparse Attention block and a TabMixer module. Their outputs are fused and aggregated across tokens before classification. The final output has shape $(\textit{batch\_size}, \textit{NUM\_CLASSES})$.}
\label{fig:model}
\end{figure*}

\begin{figure*}[!t]
\centering
\resizebox{0.6\textwidth}{!}{
    \includegraphics{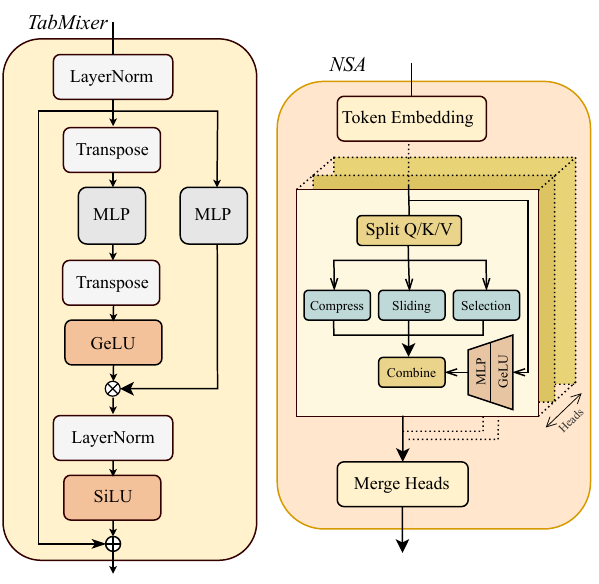}
}
\caption{TabMixer (Right) and NSA (Left) Modules}
\label{fig:modules}
\end{figure*}

Given an input tensor $X \in \mathbb{R}^{B \times N}$, where $B$ is the batch size and $N$ is the number of features, the model performs the following steps:

\begin{enumerate}
    \item \textbf{Feature Embedding:} Each scalar input is expanded into a high-dimensional representation via a shared linear embedding layer:
    \[
    X \leftarrow \texttt{Linear}(1 \rightarrow D)
    \]
    where $D$ is the embedding dimension.
     
    \item \textbf{Sparse Attention:} The embedded inputs are passed through a \textit{NSA} module. This module applies a combination of local attention (via sliding windows), block-wise compression, and block selection to reduce the quadratic complexity typically associated with full attention mechanisms.

    \item \textbf{TabMixer Module:} Parallel to the attention mechanism, the embedded inputs are also processed through a \textit{TabMixer} block, which applies channel- and token-wise MLP transformations with non-linear activations (e.g., GeLU and SiLU). This block captures complex interactions between features and across examples.

    \item \textbf{Fusion and Aggregation:} The outputs from the attention and TabMixer modules are summed element-wise, and mean pooling is applied across the token (feature) dimension:
        \[
        X \leftarrow \texttt{MeanPooling}(Y_{\text{attn}} + Z_{\text{mixer}})
        \]

    \item \textbf{Prediction Head:} A two-layer MLP with a GeLU activation reduces the feature dimension and produces the final classification logits corresponding to the number of output classes:
        \[
        output = \texttt{MLP}(X)
        \]
\end{enumerate}

This architecture combines the global context modeling benefits of attention with the efficient token-feature mixing of MLP-based architectures, enabling accurate and scalable learning from tabular data while preserving both global and local context awareness.

\subsection{TabNSA Architecture for Few Shot Learning}

We propose TabNSA+Gemma, a hybrid model that integrates pretrained large language models with the TabNSA module to enhance tabular data classification. The architecture is designed to leverage the rich semantic representations of LLMs and adapt them to structured data classification tasks through attention-based modeling through the TabNSA module. This integration is aimed at enabling effective few-shot learning by utilizing the generalization capabilities of LLMs when labeled tabular data is limited.

\subsubsection{TabNSA+Gemma}

We leverage the instruction-tuned Gemma-1.1-2B model to generate meaningful representations from natural language prompts derived from tabular features. Specifically, our architecture integrates a pretrained lightweight Gemma language model (LLM) as a feature encoder with the specialized TabNSA module for downstream classification. The LLM encoder processes tokenized input sequences to produce contextualized token embeddings. Overall, the proposed architecture comprises three primary components: a partially frozen LLM encoder, a linear adapter layer, and the TabNSA model.

To balance computational efficiency and task-specific adaptation, we freeze most parameters of the Gemma encoder, selectively fine-tuning only the final two Transformer layers. This strategy preserves the general semantic knowledge acquired during pretraining while substantially reducing computational costs and memory usage. 
These embeddings undergo a dual-pooling mechanism, consisting of both max pooling and masked mean pooling, to derive compact representations. The max-pooled embedding captures salient features, while the mean-pooled embedding summarizes the overall context. A learnable linear adapter then projects the max-pooled embedding to match the dimensionality required by TabNSA, after which it is fused with the mean-pooled embedding via element-wise addition. This combined vector, encapsulating both prominent and summary information, serves as input to the TabNSA model for final classification. Hyperparameters used for TabNSA were determined through supervised learning experiments detailed in Section~\ref{supervised:binary}.


Let $\mathbf{H} \in \mathbb{R}^{B \times T \times D}$ denote the final hidden states from the LLM for a batch of $B$ samples, each with $T$ tokens length and embedding dimension $D$. Given an attention mask $\mathbf{M} \in \{0,1\}^{B \times T}$, we compute the pooled representations:

\begin{align}
    \mathbf{h}_{\text{max}} &= \max_{t} \left( \mathbf{H}_{:, t, :} \odot \mathbf{M}_{:, t, :} + (1 - \mathbf{M}_{:, t, :}) \cdot (-\infty) \right) \label{eq:max_pool} \\
    \mathbf{h}_{\text{mean}} &= \frac{1}{\sum_{t=1}^T \mathbf{M}_{:, t}} \sum_{t=1}^T \mathbf{H}_{:, t, :} \odot \mathbf{M}_{:, t, :} \label{eq:mean_pool}
\end{align}

\noindent where $\mathbf{M}_{:, t, :} = \mathbf{M}_{:, t} \mathbf{1}_D^T$ broadcasts the mask along the embedding dimension. These are combined via:

\begin{equation}
    \mathbf{h}_{\text{adapted}} = \mathbf{W} \mathbf{h}_{\text{max}} + \mathbf{b} + \mathbf{h}_{\text{mean}} \label{eq:combined}
\end{equation}

\noindent with $\mathbf{W} \in \mathbb{R}^{D \times D}$, $\mathbf{b} \in \mathbb{R}^{D}$, yielding $\mathbf{h}_{\text{adapted}} \in \mathbb{R}^{B \times D}$ for TabNSA's input dimension $D$. The final output is:

\begin{equation}
    \mathbf{y} = \text{TabNSA}(\mathbf{h}_{\text{adapted}}) \label{eq:output}
\end{equation}

The resulting vector is then passed to the TabNSA model, which applies a combination of sparse attention and token-feature mixing to produce the final classification output. 
In our experiments, we adopt an $K$-shot training framework. 

\begin{figure*}[!t]
\centering
\includegraphics[width=\linewidth]{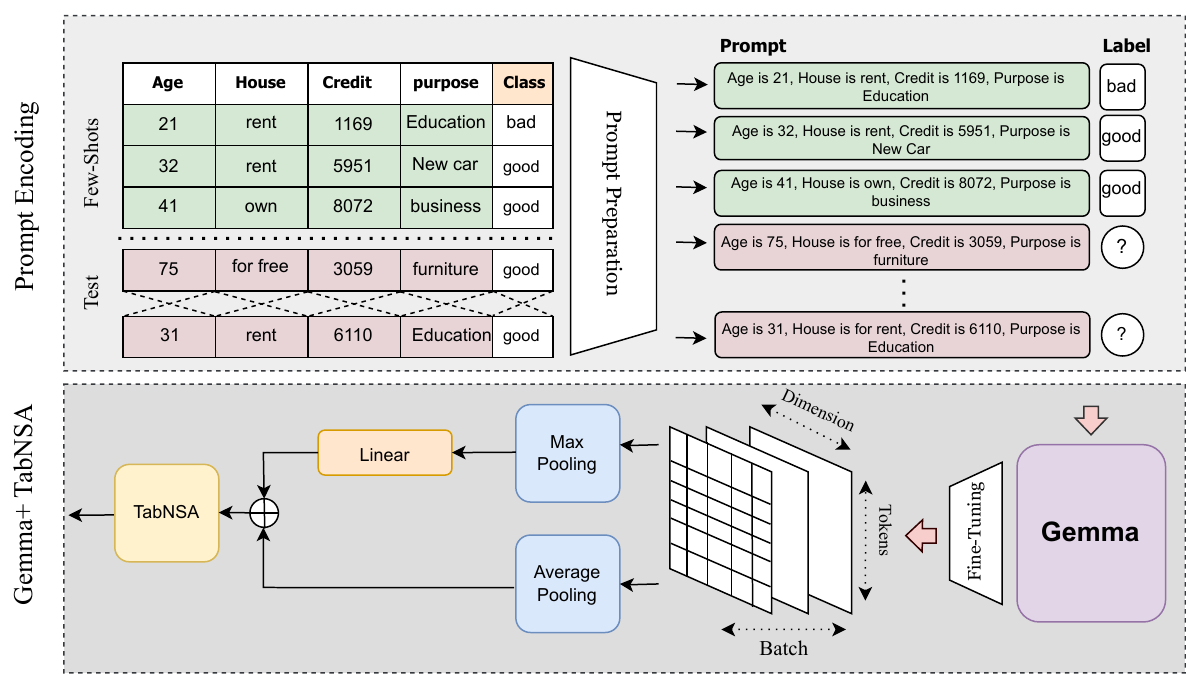}
\caption{TabNSA+Gemma for Few-Shot Learning}
\label{fig:GemmaTabNSA}
\end{figure*}

\subsubsection{Tabular-to-Text Prompt Transformation}

To enable LLM-based few-shot learning on tabular inputs, we transform each tabular row into a structured natural language prompt. Specifically, for an input instance $\mathbf{x} = [x_1, x_2, ...,$ $x_n]$ with $n$ features and label $y$, we format the prompt as:

\begin{quote}
\texttt{"Feature\_1: \{$x_1$\}, Feature\_2: \{$x_2$\}, ..., Feature\_n: \{$x_n$\}. What is the predicted label?"}
\end{quote}

For classification tasks, the model is expected to output one of the predefined label strings. During few-shot prompting, we prepend $k$ support examples (formatted similarly) to the query instance, e.g.,

\begin{quote}
\texttt{"Feature\_1: 23, Feature\_2: high, ..., Feature\_n: true. Label: Yes. \textbackslash n Feature\_1: 41, Feature\_2: low, ..., Feature\_n: false. Label: No. \textbackslash n Feature\_1: 35, Feature\_2: medium, ..., Feature\_n: true. What is the predicted label?"}
\end{quote}

Categorical features are mapped to human-readable strings, while numerical features are used verbatim, preserving original feature names to enhance interpretability. This prompt design helps the LLM leverage both structural and semantic information, aligning with recent best practices in LLM-based tabular learning. 

Furthermore, we adapt the input token length dynamically between 512 and 2048 tokens depending on the specific dataset and the typical text length associated with each entity. This strategy balances computational efficiency and context coverage, effectively utilizing the full capacity of Gemma’s attention window.

\section{Experiments and Results}

\subsection{Dataset}
In this work, we evaluated the performance of the TabNSA model and several baseline models across a diverse set of datasets. The assessment included 8 widely used binary classification datasets and 2 multi-class datasets sourced from the UCI repository, OpenML, and Kaggle. These datasets span a wide range of domains, including finance, business, chemistry, geography, ecology, image recognition, and sports, ensuring a comprehensive evaluation of the models. This diverse benchmark dataset collection enables us to systematically assess the performance across various domains and data structures. A detailed description of each dataset is provided in Table \ref{dataset}.

Experiments involving supervised learning and transfer learning were conducted on a local machine with an AMD Ryzen Threadripper PRO 5965WX 24-core CPU, 62\,GB of RAM, and an NVIDIA RTX A4500 GPU with 20\,GB of memory, as well as on P4V16 GPU nodes and NVIDIA H100-80 GPUs for the TabNSA+Gemma experiments.

\begin{table*}[ht]
    \centering
    \caption{Dataset details including abbreviation, number of classes, number of data points, and number of features.} 
    \label{dataset}
    \begin{tabular}{l c c c c}
        \toprule
        \textbf{Dataset Name} & \textbf{Abbreviation} & \textbf{\# Class} & \textbf{\# Data} & \textbf{\# Features} \\
        \midrule
        Credit-g                & CG  & 2  & 1,000 & 20 \\
        Credit-Approval         & CA  & 2  & 690 & 15 \\
        Dataset-Sales         & DS & 2  & 690 & 15 \\
        Adult         & AD  & 2  & 48,842 & 14 \\ \midrule
        Cylinder-Bands         & CB  & 2  & 540 & 35 \\
        Blastchar              & BL  & 2  & 7,043 & 35 \\
        Insurance-Co              & IO  & 2  & 5,822 & 85 \\
        1995-Income              & IC  & 2  & 32,561 & 14 \\ \midrule
        Bank              & BA  & 2  & 45,211 & 16 \\
        Blood              & BO  & 2  & 748 & 4 \\
        Diabetes              & DA  & 2  & 768 & 8 \\
        Heart              & HA  & 2  & 918 & 11 \\
        Jungle              & JU  & 2  & 44,819 & 6 \\ \midrule      
        ImageSegmentation        & SG  & 7  & 2,310 & 20 \\
        ForestCovertype          & FO  & 7  & 581,012 & 55 \\ \bottomrule
    \end{tabular}
\end{table*}


\subsection{Preprocessing}
In order to handle missing values and class imbalance we used the preprocessing method in \cite{eslamian2025tabmixer}. We used this setting for Supervised and transfer learning task. Let the input variable space as $\mathcal{D}ata\in\{\mathbb{R} \cup \mathbb{C} \cup \mathbb{B} \cup \varnothing\}$, where $\mathbb{R}$, $\mathbb{C}$, $\mathbb{B}$ and $\varnothing$ represent the domains of numerical, categorical, binary dat and NULL values respectively. After preprocessing block, we have $\{\mathcal{X}, \mathcal{Y} \}$ denote a feature-target pair, where $x$ represents numerical features, and $y$ is a continuous variable for regression tasks or a whole number for classification tasks.
\vskip 12 pt
\subsection{Experiment Set up} \label{experiment}
To enhance TabNSA's performance and stability, we employed OPTUNA \cite{optuna} to systematically tune critical hyperparameters through an efficient and automated search. Given the variability in feature distributions and dataset characteristics typical of tabular data, hyperparameter optimization is essential. Each dataset was divided into training (70\%), validation (10\%), and test (20\%) sets, with the training and validation subsets used to guide the optimization process.

In our experimental setup, we employed a two-stage optimization strategy. In the first stage, we utilized the OPTUNA package to automate the hyperparameter search specifically targeting the architecture of the NSA model. The search space included architectural parameters such as head dimension (8-46), number of attention heads (1-8), sliding window size (1-8), compression block size (4-16), selection block size (2-compression block size), and number of selected blocks (1-4). Furthermore, we fine-tuned optimization parameters like learning rate (1e-4 - 1e-3) and batch size (32-128) to strike a balance between convergence speed and generalization. 

In the second stage, having fixed the optimal architectural configuration, we trained the final model's MLP layers using the AdamW optimizer with a CrossEntropy loss function, finely tuning the learning rate (\(1 \times 10^{-4}\) to \(1 \times 10^{-3}\)) and batch size (32–128) to balance convergence speed and generalization. The model was first transferred to the designated device (CPU or GPU) and then trained in mini-batches: each batch underwent a forward pass to compute predictions, followed by loss calculation, backpropagation, and parameter updates. If a validation loader was provided, the model was evaluated at the end of each epoch without gradient tracking, and the average validation loss was monitored to implement early stopping---halting training when no improvement in validation loss was observed for a predefined number of epochs and restoring the best-performing weights. Finally, the model's performance was assessed on the test set using standard metrics, and a history of training and validation losses was returned to facilitate analysis of convergence and generalization.

For each trial, Optuna suggests values for both the architectural and optimization hyperparameters of the NSA model. This first-stage optimization integrates the full training procedure, including second-stage weight updates, to identify the hyperparameter configuration that maximizes validation AUC. Once the optimal parameters are identified, the final model is retrained on the training set and evaluated on the test set using standard metrics to assess generalization performance.

\subsection{Binary Classification} \label{supervised:binary}

We evaluated TabNSA on eight datasets for supervised binary classification, with results summarized in Table~\ref{table:binary}. Baseline results from other methods are taken directly from \cite{transtab}. Following their evaluation protocol, we ran TabNSA 10 times with different seeds and reported the average performance. TabNSA consistently outperformed classical machine learning methods, deep neural networks (DNNs), Transformer-based architectures, and tree-based models.
Although it may not achieve the best score on every dataset, TabNSA ranked first overall, demonstrating strong and consistent performance.

To assess the contribution of its components, we compare TabNSA with the NSA module alone and with TabMixer in its default configuration from \cite{eslamian2025tabmixer}. While NSA alone achieves notable improvements, TabNSA achieves a lower average rank (1.62 vs. 2.88) and a smaller standard deviation (0.744 vs. 3.36), indicating superior overall performance and greater stability across datasets.

\begin{table*}[htbp]
\centering
\caption{Performance Comparison of Models for Binary Classification} \label{table:binary}
\resizebox{\textwidth}{!}{
\begin{tabular}{lcccccccc|c}
\hline
Methods & CG & CA & DS & AD & CB & BL & IO & IC & Rank (Std) $\downarrow$ \\
\hline
Logistic Regression & 0.720 & 0.836 & 0.557 & 0.851 & 0.748 & 0.801 & 0.769 & 0.860 & 12.5 (2.33) \\
XGBoost & 0.726 & 0.895 & 0.587 & 0.912 & 0.892 & 0.821 & 0.758 & 0.925 & 7.69 (4.05) \\
MLP & 0.643 & 0.832 & 0.568 & 0.904 & 0.613 & 0.832 & 0.779 & 0.893 & 11.9 (2.01)  \\
SNN & 0.641 & 0.880 & 0.540 & 0.902 & 0.621 & 0.834 & 0.794 & 0.892 & 10.8 (3.66) \\ \midrule
TabNet & 0.585 & 0.800 & 0.478 & 0.904 & 0.680 & 0.819 & 0.742 & 0.896 & 13.7 (1.67) \\
DCN & 0.739 & 0.870 & 0.674 & 0.913 & 0.848 & 0.840 & 0.768 & 0.915 & 6.5 (2.41) \\
AutoInt & 0.744 & 0.866 & 0.672 & 0.913 & 0.808 & 0.844 & 0.762 & 0.916 & 6.5 (2.83) \\
TabTrans & 0.718 & 0.860 & 0.648 & 0.914 & 0.855 & 0.820 & 0.794 & 0.882 & 8.56 (4.12) \\ \midrule
FT-Trans & 0.739 & 0.859 & 0.657 & 0.913 & 0.862 & 0.841 & 0.793 & 0.915 & 6.5 (2.46) \\
VIME & 0.735 & 0.852 & 0.485 & 0.912 & 0.769 & 0.837 & 0.786 & 0.908 & 9.31 (2.55) \\
SCARF & 0.733 & 0.861 & 0.663 & 0.911 & 0.719 & 0.833 & 0.758 & 0.919 & 8.88 (2.89) \\
TransTab & 0.768 & 0.881 & 0.643 & 0.907 & 0.851 & 0.845 & 0.822 & 0.919 & 5.56 (2.61) \\
TabMixer \dag & 0.660 & 0.907 & 0.659 & 0.900 & 0.829 & 0.821 & 0.974 & \textbf{0.969} & 7.06 (4.96) \\ \midrule


\textbf{NSA} & \textbf{0.874} & 0.898 & \textbf{0.717} & 0.967 & 0.721 & \textbf{0.926} & 0.979 & 0.967 & 2.88 (3.36) \\

\textbf{TabNSA} & 0.859 & \textbf{0.908} & {0.690} & \textbf{0.973} & \textbf{0.898} & 0.918 & \textbf{0.984} & 0.959 & \textbf{1.62} (0.744) \\

\midrule
\end{tabular}
}

\begin{tablenotes}[flushleft]
    \footnotesize 
     \item[]  \dag Results for TabMixer correspond to the default model.
\end{tablenotes}

\end{table*}

TabNSA’s high AUC on CB (0.898) stems from its sparse attention, which mitigates overfitting in small datasets. For DS (0.690), its lower performance may reflect its complex feature interactions, potentially requiring further tuning of NSA’s selection block.

\subsection{Multi-class Classification}

Table \ref{table:multi} presents a performance comparison of TabNSA and other neural network models on two multi-class classification datasets. Since multi-class classification tasks often suffer from class imbalance, we used macro-F1 as the primary evaluation metric during training to ensure a balanced assessment across all classes, following \cite{TabKANet}. The results show that TabNSA consistently outperforms all models on both datasets.

\begin{table}[t]
    \centering
    \caption{Comparison of different methods on SG and FO datasets.}
    \label{table:multi}
    \scriptsize 
    \setlength{\tabcolsep}{3pt} 
    \begin{tabular}{l|l|cccc}
        \toprule
        Dataset & Metrics & \multicolumn{4}{c}{Method} \\
        \cmidrule(lr){3-6}
        & & MLP & TabTrans \dag & TabNet & TabNSA \\
        \midrule
        \multirow{3}{*}{SG} & ACC & 90.97 & - & 96.09 & \textbf{98.27} \\
        & F1 & 90.73 & - & 94.96 & \textbf{98.25} \\
        
        \midrule
        \multirow{2}{*}{FO} & ACC & 67.09 & 68.76 & 65.09 & \textbf{92.89} \\
        & F1 & 48.03 & 49.47 & 52.52 & \textbf{82.66} \\
        \bottomrule
    \end{tabular}

\begin{tablenotes}[flushleft]
    \footnotesize 
     \item[] \dag The SG dataset consists solely of numerical features, making it incompatible with TabTrans \cite{TabKANet}.
\end{tablenotes}
    
\end{table}

\subsection{Transfer Learning} \label{transferLearning}

Transfer learning enables machine learning models to utilize knowledge acquired from a source task to improve performance on related target tasks. While effective in domains such as computer vision and natural language processing, where structured data supports transferable representations, applying transfer learning to tabular data presents unique challenges. Unlike pixels or word embeddings with uniform structural properties, tabular data often contain heterogeneous numerical, categorical, and ordinal features lacking inherent spatial or sequential order. This feature heterogeneity can lead models trained in supervised pretraining to become overly specialized, resulting in representations that do not generalize well across tasks with differing feature compositions, class distributions, or learning objectives.

Moreover, dataset-specific feature distributions in tabular contexts frequently introduce substantial domain shifts, further limiting the adaptability of pretrained models \cite{zhang2023generative} \cite{tabtransformer}. Unlike NLP models, which benefit from common embeddings and structural regularities facilitating knowledge reuse, tabular tasks often lack such shared representations, making feature interactions context-dependent and difficult to transfer \cite{lounici2021muddling} \cite{borisov2022deep}.

To evaluate transfer learning in this context, we adopt the protocol proposed in \cite{transtab}, which introduces a controlled domain shift through partitioning each dataset into two subsets, Set1 and Set2, with approximately 50\% feature overlap. Results for baseline methods are reported from \cite{eslamian2025tabmixer}. This partial overlap creates a challenging scenario where some features are shared between domains, while others are domain-specific. This setup assesses the model’s ability to generalize common feature representations and adapt to novel, previously unseen features.

The transfer learning procedure consists of two stages: pretraining and fine-tuning. In the pretraining stage, TabNSA is trained on Set1 using supervised learning. 
After convergence on the source dataset (Set1), we conduct hyperparameter tuning specifically for the NSA module using Optuna to identify the optimal configuration. These hyperparameters are determined based on performance on a validation split of Set1. Subsequently, the best NSA hyperparameters are applied to initialize the TabNSA model, which is then trained directly on the target dataset (Set2). 
To enhance adaptation, we applied second-stage optimization for fine-tuning on Set-2. This approach enables the model to leverage insights from the source task while effectively adapting to distributional shifts and structural differences in the target domain.

By focusing attention on the most informative feature subsets during both training phases, TabNSA effectively retains transferable patterns and reduces the risk of overfitting to source-specific noise. Furthermore, the model's ability to capture both global context and local dependencies allows it to balance representational stability and adaptability under domain shifts.

We validate robustness by evaluating model performance on the test set of Set2, then repeating the procedure with reversed roles for Set1 and Set2. This bidirectional evaluation provides a comprehensive assessment of TabNSA’s transferability, highlighting its capability to maintain high performance across structurally divergent tabular tasks. The results are provided in Table \ref{table:transfer}.

\begin{table*}[htbp]

  \centering
  \caption{Evaluation of Models for Transfer Learning} \label{table:transfer}
  \resizebox{\textwidth}{!}{
  \begin{tabular}{lcccccccccccccccc|c}
    \toprule
    \multirow{2}{*}{Methods} & \multicolumn{2}{c}{CG} & \multicolumn{2}{c}{CA} & \multicolumn{2}{c}{DS} & \multicolumn{2}{c}{AD} & \multicolumn{2}{c}{CB} & \multicolumn{2}{c}{BL} & \multicolumn{2}{c}{IO} & \multicolumn{2}{c}{IC} & \multicolumn{1}{c}{Rank(Std)  $\downarrow$ } \\ 
    & set1 & set2 & set1 & set2 & set1 & set2 & set1 & set2 & set1 & set2 & set1 & set2 & set1 & set2 & set1 & set2 \\ \midrule
    Logistic Regression & 0.69 & 0.69 & 0.81 & 0.82 & 0.47 & 0.56 & 0.81 & 0.81 & 0.68 & 0.78 & 0.77 & 0.82 & 0.71 & 0.81 & 0.81 & 0.84 & 10.8 (2.86) \\
    XGBoost & 0.72 & 0.71 & 0.85 & 0.87 & 0.46 & 0.63 & 0.88 & 0.89 & \underline{0.80} & \underline{0.81} & 0.76 & 0.82 & 0.65 & 0.74 & \textbf{0.92} & 0.91 & 7.03 (3.64)  \\
    
    MLP & 0.67 & 0.70 & 0.82 & 0.86 & 0.53 & \underline{0.67} & 0.89 & \underline{0.90} & 0.73 & \textbf{0.82} & 0.79 & 0.83 & 0.70 & 0.78 & 0.90 & 0.90 & 7.25 (2.48)  \\ 
    SNN & 0.66 & 0.63 & 0.85 & 0.83 & 0.54 & 0.42 & 0.87 & 0.88 & 0.57 & 0.54 & 0.77 & 0.82 & 0.69 & 0.78 & 0.87 & 0.88 & 10.8 (3.03)  \\ \midrule 
    TabNet & 0.60 & 0.47 & 0.66 & 0.68 & 0.54 & 0.53 & 0.87 & 0.88 & 0.58 & 0.62 & 0.75 & 0.83 & 0.62 & 0.71 & 0.88 & 0.89 & 11.9 (2.47)   \\
    DCN & 0.69 & 0.70 & 0.83 & 0.85 & 0.51 & 0.58 & 0.88 & 0.74 & 0.79 & 0.78 & 0.79 & 0.76 & 0.70 & 0.71 & 0.91 & 0.90 & 8.34 (2.95)  \\
    AutoInt & 0.70 & 0.70 & 0.82 & 0.86 & 0.49 & 0.55 & 0.88 & 0.74 & 0.77 & 0.79 & 0.79 & 0.76 & 0.71 & 0.72 & 0.91 & 0.90 & 8.38 (2.92)  \\ 
    TabTrans & 0.72 & 0.72 & 0.84 & 0.86 & 0.54 & 0.57 & 0.88 & \underline{0.90} & 0.73 & 0.79 & 0.78 & 0.81 & 0.67 & 0.71 & 0.88 & 0.88 & 8.47 (2.88)  \\  \midrule
    FT-Trans & 0.72 & 0.71 & 0.83 & 0.85 & 0.53 & 0.64 & 0.89 & 0.90 & 0.76 & 0.79 & 0.78 & 0.84 & 0.68 & 0.78 & 0.91 & 0.91 & 6.41 (2.03)  \\ 
    VIME & 0.59 & 0.70 & 0.79 & 0.76 & 0.45 & 0.53 & 0.88 & \underline{0.90} & 0.65 & \underline{0.81} & 0.58 & 0.83 & 0.67 & 0.70 & 0.90 & 0.90 & 10.9 (3.04)  \\
    SCARF & 0.69 & 0.72 & 0.82 & 0.85 & 0.55 & 0.64 & 0.88 & 0.89 & 0.77 & 0.73 & 0.78 & 0.83 & 0.71 & 0.75 & 0.90 & 0.89 & 7.31 (2.07)  \\ 
    
    TransTab & 0.74 & 0.76 & 0.87 & 0.89 & 0.55 & 0.66 & 0.88 & \underline{0.90} & \underline{0.80} & 0.80 & 0.79 & 0.84 & 0.73 & 0.82 & 0.91 & 0.91 & 3.69 (1.47)  \\ 

    TabMixer \dag & \underline{0.86} & \underline{0.84} & \underline{0.87} & \underline{0.88} & 0.64 & \textbf{0.71} & \underline{0.90} & \underline{0.90} & \textbf{0.94} & 0.77 & \textbf{0.93} & \textbf{0.92} & \underline{0.95} & \underline{0.95} & \textbf{0.94} & \textbf{0.95} & \underline{1.91} (0.89)  \\ \midrule

    \textbf{TabNSA} & \textbf{0.91} & \textbf{0.90} & \textbf{0.93} & \textbf{0.94} & \textbf{0.68} & \textbf{0.71} & \textbf{0.96} & \textbf{0.95} & 0.79 & 0.76 & \underline{0.92} & \textbf{0.92} & \textbf{0.98} & \textbf{0.98} & \textbf{0.94} & \underline{0.94} & \textbf{1.84} (1.75) \\  \bottomrule

  \end{tabular} 
  }

\begin{tablenotes}[flushleft]
\footnotesize 
 \item[]  \dag Results for TabMixer correspond to the default model.
\end{tablenotes}

\end{table*}


\subsection{Few-Shot Learning} \label{Few-Shot}

Few-shot learning (FSL) enables models to generalize effectively from limited labeled data, a critical challenge in tabular data applications where labeled examples may be scarce due to cost, privacy, or class imbalance. The results for few-shot learning using TabNSA+Gemma is reported in \ref{table:few-shot}. We used the results of baseline methods from \cite{tabllm}.


\begin{table*}[t]
\caption{Evaluation of Models for Sample Few-Shot Learning}
\centering
\setlength{\tabcolsep}{8pt}
\resizebox{\textwidth}{!}{\begin{tabular}{llcccccccc}
\toprule
& &\multicolumn{8}{c}{\textbf{Number of Shots}}\\
\addlinespace[1.5mm]
\textbf{Dataset} & \textbf{Method} & \textbf{4} & \textbf{8} & \textbf{16} & \textbf{32} & \textbf{64} & \textbf{128} & \textbf{256} & \textbf{512}\\
\midrule
\multirow{4}{*}{BA} 
& XGBoost  & $0.50_{.00}$   & $0.56_{.09}$   & $0.68_{.04}$   & $\mathbf{0.76}_{.03}$   & $\mathbf{0.83}_{.02}$   & $0.85_{.03}$   & $0.88_{.01}$   & $0.90_{.01}$  \\
& TabPFN   & $0.59_{.14}$   & $\mathbf{0.66}_{.08}$   & $\mathbf{0.69}_{.02}$   & $\mathbf{0.76}_{.03}$   & $0.82_{.03}$   & $\mathbf{0.86}_{.02}$   & $\mathbf{0.89}_{.00}$   & $\mathbf{0.90}_{.00}$     \\
& TabLLM       & $\mathbf{0.59}_{.10}$    & $0.64_{.05}$    & $0.65_{.05}$    & $0.64_{.06}$    & $0.69_{.03}$    & $0.82_{.05}$    & $0.87_{.01}$    & $0.88_{.01}$ \\
& TabNSA+Gemma     & $0.56_{.30}$    & $0.45_{.05}$    & $0.57_{.02}$  & $0.60_{.01}$ & $0.65_{.04}$ & $0.69_{.04}$ & $0.67_{.01}$ & $0.75_{.01}$ \\
\midrule
\multirow{3}{*}{BO} 
& XGBoost                                    & $0.50_{.00}$   & $0.58_{.07}$   & $0.66_{.04}$   & $0.67_{.06}$   & $0.68_{.05}$   & $0.71_{.06}$   & $0.70_{.07}$   & $0.67_{.06}$  \\
& TabPFN                                    & $0.52_{.08}$   & $0.64_{.04}$   & $\mathbf{0.67}_{.01}$   & $0.70_{.04}$   & $0.73_{.04}$   & $0.75_{.04}$   & $0.76_{.04}$   & $0.76_{.03}$  \\
& TabLLM                          & $0.58_{.09}$    & $\mathbf{0.66}_{.03}$    & $0.66_{.07}$    & $0.68_{.04}$    & $0.68_{.04}$    & $0.68_{.06}$    & $0.70_{.08}$    & $0.68_{.04}$  \\
& TabNSA+Gemma      & $\mathbf{0.61}_{.08}$    & $0.61_{.02}$    & $0.62_{.03}$  & $\mathbf{0.81}_{.03}$  & $\mathbf{0.82}_{.01}$ & $\mathbf{0.87}_{.04}$ & $\mathbf{0.87}_{.04}$ & $\mathbf{0.87}_{.05}$ \\
\midrule
\multirow{4}{*}{CG} 
& XGBoost                                   & $0.50_{.00}$   & $0.51_{.07}$   & $0.59_{.05}$   & $0.66_{.03}$   & $0.67_{.06}$   & $0.68_{.02}$   & $0.73_{.02}$   & $0.75_{.03}$  \\
& TabPFN                                      & $0.58_{.08}$   & $0.59_{.03}$   & $0.64_{.06}$   & $0.69_{.07}$   & $0.70_{.07}$   & $0.72_{.06}$   & $\mathbf{0.75}_{.04}$   & $0.75_{.02}$    \\
& TabLLM & $\mathbf{0.69}_{.04}$    & $\mathbf{0.66}_{.04}$    & $0.66_{.05}$    & $0.72_{.06}$    & $0.70_{.07}$    & $0.71_{.07}$    & $0.72_{.03}$    & $0.72_{.02}$ \\
& TabNSA+Gemma                             & $0.44_{.02}$    & $0.64_{.07}$ & $\mathbf{0.71}_{.01}$    & $\mathbf{0.72}_{.01}$    & $\mathbf{0.71}_{.02}$ & $\mathbf{0.74}_{.05}$ & $0.75_{.40}$ & $\mathbf{0.79}_{.03}$ \\
\midrule
\multirow{3}{*}{DA} 
& XGBoost                                 & $0.50_{.00}$   & $0.59_{.16}$   & $\mathbf{0.72}_{.07}$   & $0.69_{.08}$   & $0.73_{.05}$   & $0.78_{.05}$   & $0.80_{.03}$   & $0.80_{.01}$         \\
& TabPFN                                      & $0.61_{.13}$   & $\mathbf{0.67}_{.11}$   & $0.71_{.07}$   & $\mathbf{0.77}_{.03}$   & $\mathbf{0.82}_{.03}$   & $\mathbf{0.83}_{.03}$   & $\mathbf{0.83}_{.03}$   & $\mathbf{0.81}_{.02}$  \\
& TabLLM                           & $\mathbf{0.61}_{.09}$    & $0.63_{.08}$    & $0.69_{.07}$    & $0.68_{.04}$    & $0.73_{.03}$    & $0.79_{.04}$    & $0.78_{.02}$    & $0.78_{.04}$         \\
& TabNSA+Gemma      & $0.55_{.18}$ & $0.65_{.14}$ & $0.69_{.01}$    & $0.60_{.01}$    & $0.70_{.02}$ & $0.76_{.01}$ & $0.79_{.01}$ & $\mathbf{0.81}_{.02}$ \\
\midrule
\multirow{3}{*}{HA} 
& XGBoost                                  & $0.50_{.00}$   & $0.55_{.14}$   & $0.84_{.07}$   & $0.88_{.04}$   & $0.91_{.01}$   & $0.91_{.01}$   & $0.90_{.01}$   & $\mathbf{0.92}_{.01}$            \\
& TabPFN     & $0.84_{.06}$   & $\mathbf{0.88}_{.05}$   & $0.87_{.06}$   & $0.91_{.02}$   & $\mathbf{0.92}_{.02}$   & $\mathbf{0.92}_{.02}$   & $0.92_{.01}$   & $0.92_{.02}$  \\
& TabLLM                          & $0.76_{.14}$    & $0.83_{.05}$    & $0.87_{.04}$    & $0.87_{.06}$    & $0.91_{.01}$    & $0.90_{.01}$    & $\mathbf{0.92}_{.01}$    & $\mathbf{0.92}_{.01}$      \\
& TabNSA+Gemma      & $\mathbf{0.87}_{.03}$ & $0.87_{.03}$  & $\mathbf{0.88}_{.01}$  & $\mathbf{0.92}_{.04}$  & $0.91_{.02}$ & $0.91_{.02}$ & $\mathbf{0.92}_{.01}$ & $0.90_{.02}$ \\
\midrule
\multirow{3}{*}{AD} 
& XGBoost  & $0.50_{.00}$   & $0.59_{.06}$   & $0.77_{.02}$   & $0.79_{.03}$   & $0.82_{.02}$   & $0.84_{.01}$   & $0.87_{.01}$   & $0.88_{.00}$  \\
& TabPFN                                    & $0.73_{.08}$   & $0.71_{.09}$   & $0.76_{.09}$   & $0.80_{.04}$   & $0.82_{.04}$   & $0.84_{.01}$   & $0.86_{.01}$   & $0.87_{.01}$ \\
& TabLLM   & $\mathbf{0.84}_{.01}$    & $\mathbf{0.84}_{.02}$    & $\mathbf{0.84}_{.04}$    & $\mathbf{0.84}_{.01}$    & $\mathbf{0.84}_{.02}$    & $\mathbf{0.86}_{.01}$    & $\mathbf{0.87}_{.00}$    & $\mathbf{0.89}_{.01}$              \\
& TabNSA+Gemma      & $0.73_{.04}$    & $0.71_{.05}$    & $0.73_{.01}$    & $0.72_{.00}$ & $0.77_{.04}$ & $0.85_{.01}$ & $0.87_{.01}$ & $0.86_{.01}$ \\
\midrule
\multirow{3}{*}{JU} 
& XGBoost                                 & $0.50_{.00}$   & $0.58_{.07}$   & $\mathbf{0.72}_{.05}$   & $0.78_{.03}$   & $0.81_{.02}$   & $0.84_{.02}$   & $0.87_{.01}$   & $0.91_{.01}$   \\
& TabPFN                                     & $0.65_{.08}$   & $\mathbf{0.72}_{.04}$   & $0.71_{.07}$   & $\mathbf{0.78}_{.02}$   & $\mathbf{0.81}_{.01}$   & $\mathbf{0.84}_{.01}$   & $\mathbf{0.88}_{.01}$   & $0.91_{.00}$   \\
& TabLLM                          & $0.64_{.01}$    & $0.64_{.02}$    & $0.65_{.03}$    & $0.71_{.02}$    & $0.78_{.02} $    & $0.81_{.02}$    & $0.84_{.01}$    & $0.89_{.01}$  \\
& TabNSA+Gemma     & $\mathbf{0.67}_{.01}$ & $0.66_{.02}$ & $0.66_{.04}$ & $0.68_{.20}$  & $0.67_{.02}$ & $0.70_{.01}$ & $0.69_{.01}$ & $\mathbf{0.94}_{.01}$ \\

\midrule
\end{tabular}}


\label{table:few-shot}
\end{table*}

By coupling semantic embeddings from large language models with sparse attention and token–feature mixing, TabNSA+Gemma excels in semi-structured, low-resource scenarios, where text-derived representations enrich structured-data interpretation. In few-shot learning experiments, It shows a clear advantage on small datasets, while still achieving competitive results on larger ones. On Blood (BO; 748 samples) it achieves 0.61 BA in the 10-shot setting and 0.81 BA in the 500-shot setting; on Credit-g (CG; 1000 samples) it leads with 0.71 BA at 100 shots, 0.72 BA at 500 shots, and 0.79 BA at 1000 shots; and on Heart (HA; 918 samples) it secures 0.87 BA at 10 shots, 0.88 BA at 100 shots, and 0.92 BA at 500 shots. This consistent, low-variance performance on datasets an order of magnitude smaller than AD (45,211 rows), BA (48,842 rows), and JU (44,819 rows) highlights TabNSA+Gemma’s robustness and noise resilience in few-shot learning environments.

\subsection{Analysis of the Performance of TabNSA}
Across the experimental results, TabNSA consistently outperforms both deep learning and ensemble baselines, particularly in settings with limited labeled data or high feature dimensionality. This performance can be attributed to TabNSA’s dynamic sparse attention, which enables instance-specific feature selection and reduces overfitting by ignoring irrelevant or redundant features. For example, in the BO dataset, where sample sizes are small and feature interactions are sparse, TabNSA’s coarse-to-fine feature selection significantly improves generalization. On transfer and incremental learning benchmarks, the model benefits from its ability to adaptively focus on newly introduced or domain-shifted features without retraining the entire architecture. Furthermore, the integration with the TabMixer backbone allows TabNSA to capture complex nonlinear dependencies among selected features, which boosts performance in more structured datasets such as Covertype or Adult. These results highlight the strength of combining sparsity-driven attention with lightweight, expressive mixing for tabular data modeling.

While TabNSA+Gemma achieves strong results on smaller datasets such as Blood (BO), Credit-g (CG), and Heart (HA), its performance lags behind baselines like TabPFN and TabLLM on larger datasets such as Bank (BA) and Dresses-Sales (DS) in low-shot scenarios (e.g., 4–16 shots). This discrepancy can be attributed to a combination of dataset complexity and limitations in prompt-based adaptation. The BA and DS datasets, with 45,211 and 690 samples respectively, exhibit high feature heterogeneity and intricate inter-feature relationships, which are challenging to capture through static prompt templates. In particular, the tabular-to-text transformation used in TabNSA+Gemma encodes feature values in a literal, context-agnostic form, which may fail to convey deeper structural patterns—especially when training examples are few and semantically diverse.

Moreover, while Gemma-1.1-2B is instruction-tuned and effective in many reasoning tasks, it may not be optimally aligned with the distributions and tokenization challenges present in larger tabular datasets. Without task-specific adaptation, the language model's internal representations may fail to generalize from a small number of token-level examples. To mitigate this, future work could explore more expressive prompt engineering strategies (e.g., incorporating per-feature importance weights, natural-language feature descriptions, or learned embeddings) and lightweight fine-tuning of Gemma’s internal layers (e.g., LoRA or prompt tuning) to better tailor it to structured data under few-shot constraints.

\subsubsection{Model Computational Efficiency}
We assess the computational efficiency of four models—TransTab, TabMixer, NSA, and our proposed TabNSA by reporting their FLOPs and parameter counts, computed using the \texttt{ptflops} library \cite{ptflops}, as well as their overall runtime measured from model initialization to the end of training on a shared hardware setup. The results ares shown in Table \ref{table:efficiency}.

\begin{table*}[htbp]
  \centering
  \caption{Comparison of the Number of FLOPs, Run Time and the Number of Parameters for CG and BL Datasets on TransTab and TabMixer} \label{table:efficiency}
  \resizebox{0.8\textwidth}{!}{
  \begin{tabular}{lcc|cc|cc}
    \toprule
    \multirow{2}{*}{Experiment} & \multicolumn{2}{c}{\# FLOPS  $\downarrow$ } & \multicolumn{2}{c}{Run Time (sec) $\downarrow$} & \multicolumn{2}{c}{\# Parameters $\downarrow$} \\ 
    & CG & BL & CG & BL & CG & BL  \\ 
    \midrule
    TransTab & 4.07B & 203.7B & 50.26 & 1076.47 & 4.08M & 28.9M \\
    NSA & 55.92M & 41.08M & 8.12 & 43.47 & 304.57k & 304.57k \\  \midrule
    
    TabNSA & 64.16M & 47.2M & 8.65 & 46.78 & 369.46k & 361.25k \\ 
    \bottomrule
  \end{tabular}
  }
\end{table*}

Although training runtime is influenced by factors such as learning rate, number of epochs, and early stopping, FLOPs and parameter count serve as hardware-agnostic indicators of model complexity and scalability.

Our findings show that NSA reduces FLOPs by approximately 99\% (from 4 billion to 55 million) while improving AUC by 13\% (from 0.768 to 0.874), highlighting the effectiveness of sparse feature interactions. TabNSA, despite only a slight increase in parameter count over NSA, achieves a better average rank score. These results demonstrate that TabNSA offers a practical balance between performance and computational efficiency, making it well-suited for tabular data modeling in resource-constrained environments.

\subsubsection{Computational Complexity Analysis}
The computational complexity of the TabNSA model is derived from its key components. We summarize the notations as follows: 
\begin{itemize}
    \item $B$: Batch size
    \item $N$: Number of features
    \item $D$: Embedding dimension
    \item $H$: Number of attention heads
    \item $D_H = D/H$: Dimension per head
    \item $w$: Sliding window size
    \item $g$: Global tokens ($g = \mathtt{num\_selected\_blocks} \times \mathtt{selection\_block\_size}$)
\end{itemize}

Based on the model definition provided in Section \ref{Model_description}, the component-wise computational complexity can be outlined for the following key modules: Feature Embedding, SparseAttention, TabMixer Block, Pooling \& Residual Connection, and the Classification Head. The complexity associated with each component is summarized in Table \ref{table:Complexity}.

\begin{table}[ht]
\centering
\caption{Component-wise Complexity of the Model} \label{table:Complexity}
\begin{tabular}{@{}ll@{}}
\toprule
\textbf{Component} & \textbf{Complexity} \\
\midrule
Feature Embedding & $O(BND)$ \\
\midrule
\multirow{3}{*}{SparseAttention} 
    & Query/Key/Value Projections: $O(BN^2D_H)$ \\
    & Attention Computation: $O(BN(w + g)D_H)$ \\
    & Output Projection: $O(BN^2D_H)$ \\
\midrule
TabMixer Block & $O(BND)$ \\
\midrule
Pooling \& Residual Connection & $O(BND)$ \\
\midrule
Classification Head & $O(BD)$ \\
\bottomrule
\end{tabular}
\end{table}
The overall complexity is dominated by the projection operations in SparseAttention:
\begin{equation}
    O(BN^2D_H) + O(BN(w + g)D_H) + O(BND) + O(BD) = O(BN^2D_H)
\end{equation}
Therefore, the asymptotic complexity of the TabNSA forward pass is $O(BN^2D_H)$.

\noindent {The key Observations are:}
\begin{itemize}
    \item The $D^2_H$ term originates from linear projections in sparse attention layers.  We assume that the number of features $N$ is greater than the sum of the local window size $w$ and the number of global tokens $g$ ($N > (w + g)$).

    \item Attention computation is linear in $N$ due to sparsity ($w$ and $g$ are constants)
    \item TabMixer contributes only lower-order terms ($O(BND)$)
    \item Complexity scales linearly with batch size and feature count
\end{itemize}

\section{Ablation Study}

\begin{figure*}[!t]
\centering
\includegraphics[width=0.7\linewidth]{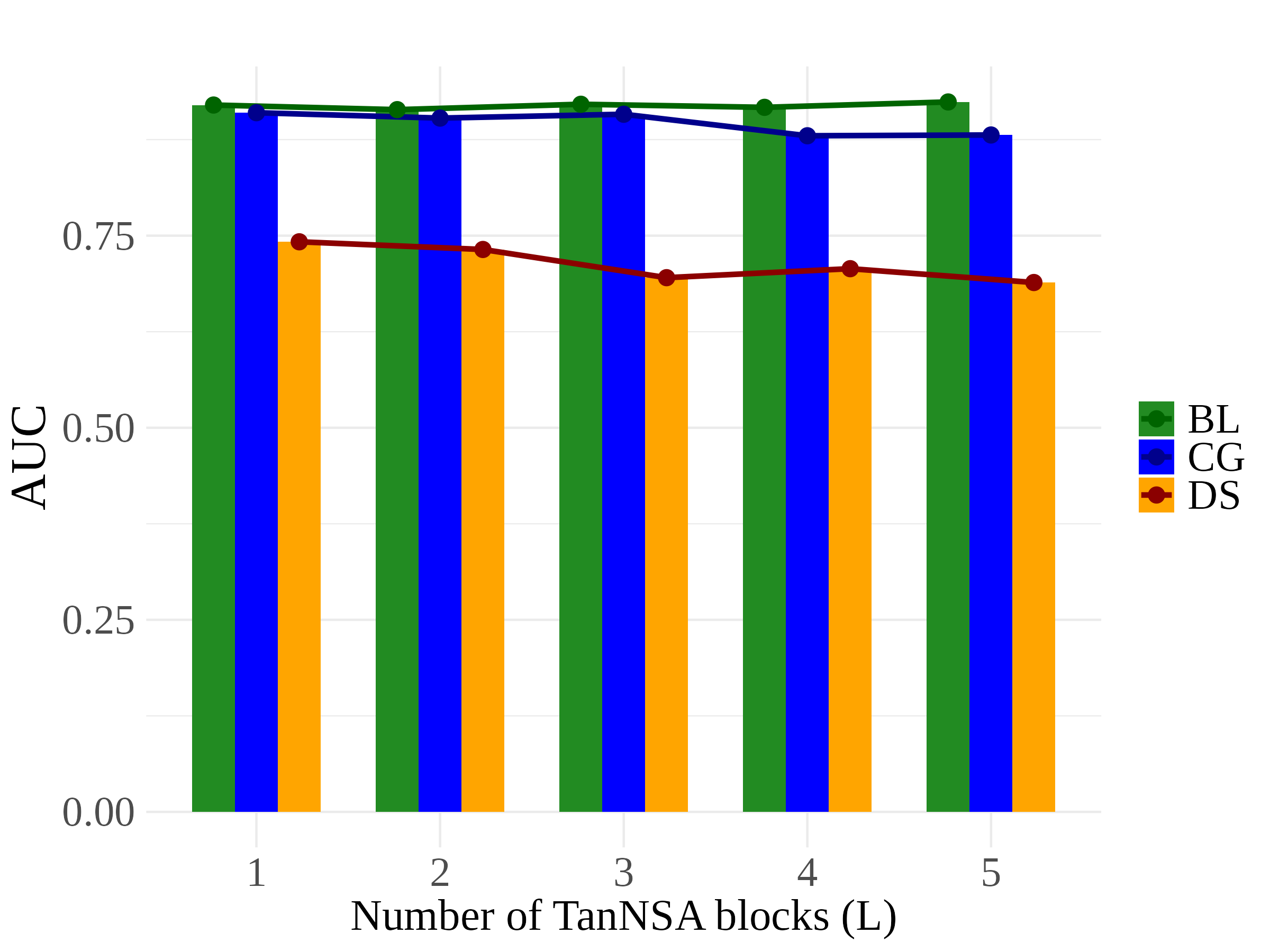}
\caption{\textbf{Impact of the Number of TabMixer Blocks on Performance}: The figure shows that varying the number of TabMixer blocks (L) has minimal impact on performance.}
\label{fig:LBlock}
\end{figure*}

\noindent  \textbf{Cascading TabNSA Blocks}.
In this study, we examine how varying the number of TabNSA blocks influences overall performance, as illustrated in Fig.\ref{fig:LBlock}. Our findings show that increasing the number of $L$ blocks does not significantly affect the final results. In most cases, using a single TabNSA block ($L=1$) is sufficient to achieve the performance levels reported in Table \ref{supervised:binary} across the benchmark datasets. This indicates that our model can deliver strong results with a relatively simple design. However, we recommend tuning this hyperparameter based on the specific characteristics of each dataset in practical applications.\\

\begin{figure*}[!t]
\centering
\caption{Ablation study of the TabNSA module under different fusion strategies. The variations include: (1) original TabNSA (TabNSA-o), (2) MLP fusion (TabNSA-m), (3) concatenation fusion (TabNSA-c), and (4) residual block-based fusion (TabNSA-r). Each configuration combines the TabMixer and NSA modules in a unique way to evaluate performance impacts.}
\resizebox{\textwidth}{!}{
    \includegraphics{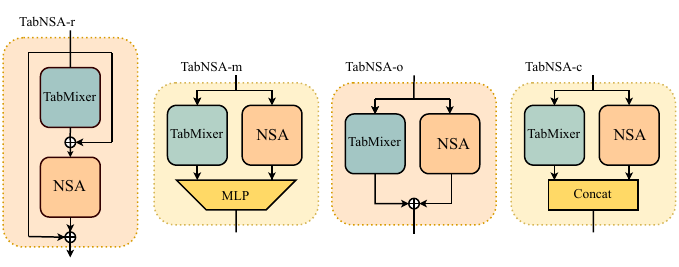}
}
\label{fig:ablation}
\end{figure*}

\noindent  \textbf{Analysis of Fusion Strategies in the Proposed Model}.
In the core idea of the TabNSA module (see Fig.\ref{fig:model}), we fused the TabMixer and NSA modules using a multiplied-addition operator. This final architecture was obtained through several experiments across different scenarios. We altered the core of the TabNSA module by considering the original version (TabNSA-o), fusing with an MLP (TabNSA-m), fusing by concatenation (TabNSA-c), and using consecutive blocks with residual connections (TabNSA-r). Fig.\ref{fig:ablation} provides a clear illustration of these different scenarios.

To validate the main architecture, we conducted experiments using two datasets for binary supervised classification. The results can be seen in Table \ref{tab:fusion}. In this experiment, we keep the hyperparameters of the NSA module the same for all datasets and run the code once with the same seed across all datasets.

\begin{table}[ht]
\centering
\caption{AUC (\%) performance of different TabNSA variants on CG and BL datasets.}
\begin{tabular}{l|ccc}
\hline
\textbf{Dataset} & \textbf{CG} & \textbf{BL} & \textbf{DS}\\
\hline
TabNSA-r  & 90.3 & 91.4 & 77.6 \\
TabNSA-m  & 89.8 & 89.9 & 73.2 \\
TabNSA-o  & \textbf{91.0} & \textbf{92.7} & 75.5 \\
TabNSA-c  & 88.4 & 91.9 & \textbf{77.9} \\
\hline
\end{tabular}
\label{tab:fusion}
\end{table}

\noindent  \textbf{Ablation on Optimizer}.
In our ablation study, for the second-stage optimization approach discussed in \ref{experiment}, we conduct ablations on two strategies: the AdamW optimizer via the standard \texttt{fit} function, and full-batch optimization using the LBFGS algorithm \cite{moritz2016linearly}, a quasi-Newton method, (\texttt{fit\_lbfgs}). The \texttt{fit} function performs optimization using mini-batch updates with shuffled data, making it efficient for larger datasets \cite{guan2023weight}. Conversely, \texttt{fit\_lbfgs} leverages a full-batch approach, which is typically more suitable for smaller datasets or scenarios where precise gradient computations per epoch are beneficial.

L-BFGS approximates second-order information by maintaining a limited history of gradient and variable updates, enabling superlinear convergence with significantly lower memory overhead than full BFGS methods. Its efficiency and strong convergence properties make it especially useful in smaller-scale or highly smooth optimization landscapes \cite{aggrawal2021hessian} \cite{gao2019quasi}. This makes L-BFGS a powerful alternative in scenarios where full-batch computation is feasible, allowing for effective exploration of loss landscapes and stable parameter updates.
The LBFGS method requires a closure function to recompute gradients during each optimization step, ensuring accurate updates based on the full batch. Comparing these two approaches allowed us to analyze trade-offs between convergence speed, computational efficiency, and final performance, and we selected the optimizer that yielded better results for each dataset as reported in Table~\ref{tab:optimizer}.

\begin{table}[ht]
\centering
\caption{Comparison of optimization strategies (AdamW vs. L-BFGS) in terms of AUC score (\%) across datasets.}
\begin{tabular}{lcc}
\hline
\textbf{Dataset} & \textbf{AdamW AUC (\%)} & \textbf{L-BFGS AUC (\%)} \\
\hline
CA  & 91.62 & 91.62 \\
DS  & 55.08 & 72.50 \\
CG  & 87.39 & 87.14 \\
\hline
\end{tabular}
\label{tab:optimizer}
\end{table}

\noindent  \textbf{Ablation Study on Pre-trained Large Language Models}.
To evaluate the impact of different pre-trained language models on overall system performance, we conducted an ablation study by substituting the core language model component with four popular large language models: BERT \cite{koroteev2021bert}, RoBERTa \cite{liu2019roberta}, QWen \cite{qwen3embedding}, and Gemma.
Each model was integrated into our architecture with minimal modifications to ensure a fair comparison. The downstream task remained consistent across experiments, allowing us to isolate the effect of the pre-trained backbone on performance.

\begin{table*}[ht]
\centering
\caption{Ablation study on different pre-trained LLMs used in our model.}
\label{tab:llm_ablation}
\resizebox{0.8\textwidth}{!}{
\begin{tabular}{lcc}
\hline
\textbf{Model} & \textbf{Parameters} & \textbf{Notes} \\
\hline
\texttt{bert-base-uncased} & 110M & Baseline encoder model \\
\texttt{roberta-base} & 125M & Improved training/data over BERT \\
\texttt{Qwen/Qwen3-Embedding-0.6B} & 600M  & Optimized for embedding tasks \\
\texttt{google/gemma-1.1-2b-it} & 2B & Instruction-tuned, strong reasoning \\
\hline
\end{tabular}}
\end{table*}

\begin{figure*}[!t]
\centering
\includegraphics[width=0.7\linewidth]{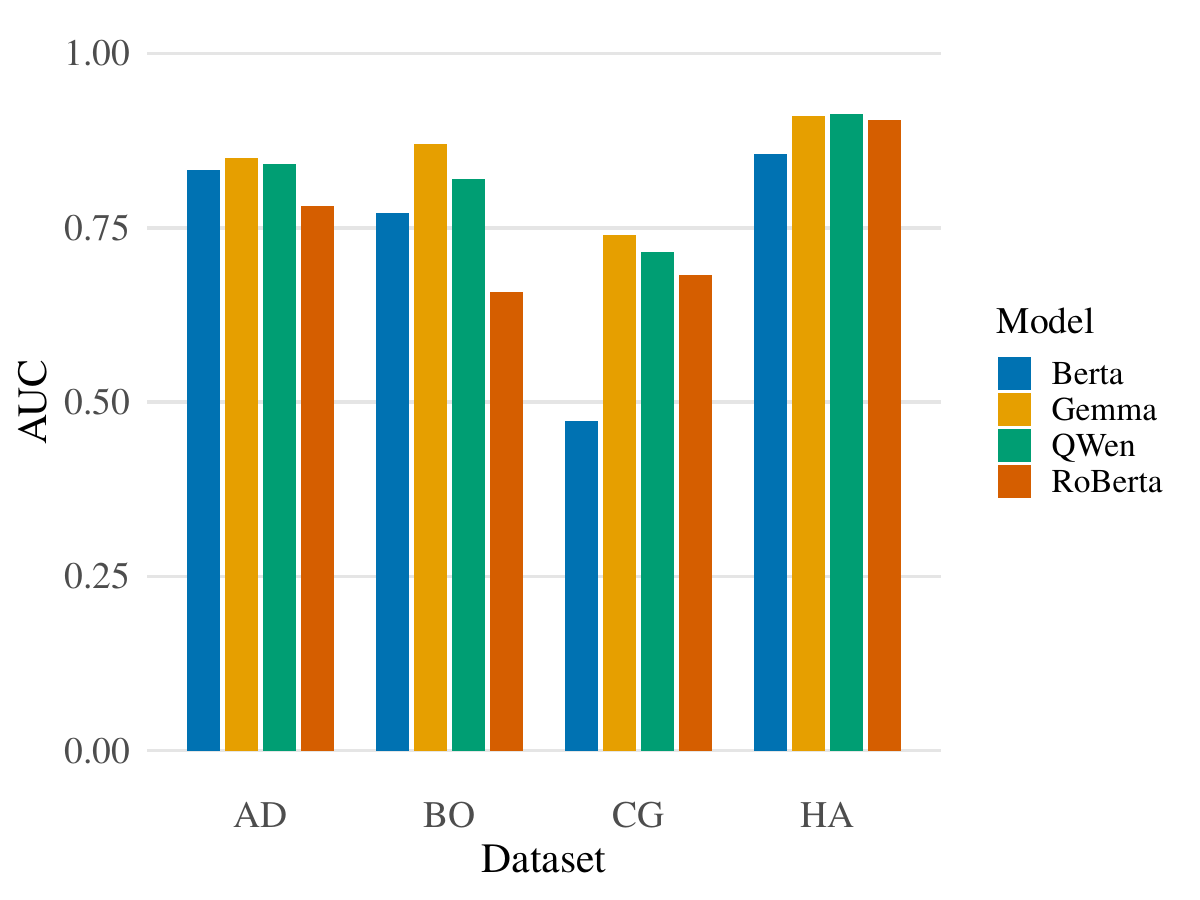}
\caption{TabNSA+Gemma for Few-Shot Learning}
\label{fig:LLM}
\end{figure*}

\noindent  \textbf{Ablation of Fine-Tuning LLM}.
In the fine-tuning process described in Section \ref{Few-Shot}, we froze all layers of Gemma and fine-tuned only the last two layers. We also conducted ablation studies using alternative fine-tuning methods, including Low-Rank Adaptation (LoRA) techniques and MixLoRA \cite{li2024mixlora}. LoRA is a Parameter-Efficient Fine-Tuning (PEFT) method. Additionally, we evaluated the performance using the original pre-trained weights of Gemma without any fine-tuning.

Due to the high complexity associated with MixLoRA and LoRA, we preferred standard fine-tuning methods, which maintain simplicity and achieve comparable or even better AUC scores. Specifically, for this task, only 5,470,213 out of 2,511,642,641 parameters (approximately 0.22\%) needed fine-tuning.

\section{Conclusion}

In this paper, we introduced TabNSA, a novel deep learning framework that adapts Native Sparse Attention (NSA) for tabular data learning. TabNSA addresses the computational inefficiencies of traditional attention mechanisms by dynamically focusing on relevant features through a hierarchical sparse strategy combining token compression, selection, and sliding windows.

Our experiments demonstrate TabNSA's superior performance on both binary and multi-class classification tasks, indicating potential for further improvement. Compared to previous Transformer-based approaches, TabNSA offers reduced computational complexity, hardware-aligned design, and inherent integration of sparsity during training. These properties enable effective balancing between global context and local feature interactions.

Tabular data presents unique challenges for deep learning due to heterogeneous feature types and the absence of inherent spatial or sequential structure. To address these challenges, we combined NSA with TabMixer. NSA reduces the quadratic complexity of standard attention mechanisms by employing hierarchical sparse strategies: Token Compression aggregates sequential blocks of keys or values; Token Selection preserves important keys and values based on importance scores; and Sliding Windows capture local patterns within fixed-size windows. This approach efficiently handles heterogeneous features by dynamically adjusting feature importance.

Complementing NSA, the TabMixer module processes information via residual connections, channel-wise mixing, and token-wise mixing using parallel MLPs with independent learning parameters and SiLU nonlinear activation functions. TabMixer effectively captures complex, non-linear global dependencies by simultaneously exploring interactions across features and examples.

The synergistic fusion of NSA and TabMixer, through element-wise summation and mean pooling, enables TabNSA to maintain global context awareness, local precision, and computational efficiency. Extensive experiments demonstrate that TabNSA consistently outperforms state-of-the-art deep learning and ensemble models across diverse benchmark datasets. The results show advantages in both supervised and transfer learning scenarios. Additionally, when the Gemma-2B model is ensembled with slight modifications to the enhanced architecture, it performs well on some datasets in few-shot learning settings. Future work could extend TabNSA to time-series tabular data by incorporating temporal attention or improve regression performance by integrating soft attention and Gaussian loss functions.

\section{Acknowledgements}
We thank the creators of the public datasets and the authors of the baseline models for making these resources available for research. We gratefully acknowledge Brian Gold, PhD, and the Gold Lab at the University of Kentucky for their support and for providing the facilities necessary to carry out this research. We gratefully acknowledge the University of Kentucky Center for Computational Sciences and Information Technology Services Research Computing for providing access to the Lipscomb Compute Cluster and related resources.  This research is supported in part by the NSF under Grant IIS 2327113 and ITE 2433190, the NIH under Grants R21AG070909 and P30AG072946, and the National Artificial Intelligence Research Resource (NAIRR) Pilot NSF OAC 240219, and Jetstream2, Bridges2, and Neocortex Resources. In particular, we used the Bridge-2 system (with NVIDIA H100-80 GPUs) and the Neocortex high-performance computing systems at the Pittsburgh Supercomputing Center to conduct the TabNSA+Gemma experiments, which required specialized hardware support.



\bibliographystyle{unsrt}

\newpage
\begin{appendices}


\section{Feature Few-Shot Learning}

For this scenario, we followed the instructions in Section \ref{transferLearning} to divide the data into two sets, with a 10\% overlap in features. In this setup, TabNSA+Gemma is first trained on the initial set to obtain the NSA parameters. Then, in the second set, we apply a second-stage optimization to fine-tune the model on the test data. This approach can be regarded as Feature-Few-Shot Learning, since the source task has no information about the target task's features. However, the same data samples are used in both subsets. This scenario is particularly relevant in healthcare applications, where new features for a subject may become available over time. It can also be viewed as a challenging case of Feature-Incremental Learning. Results are shown in Table \ref{table:feature-few-shot}.

\begin{table*}[htbp]
\centering
\caption{Evaluation of Models for Feature-Few-Shot Learning}  \label{table:feature-few-shot}
\resizebox{0.8\textwidth}{!}{
\begin{tabular}{lcccccc}
\hline
Methods & CG & CA & DS  & CB & BL  \\
\hline
BertTabNSA & 0.584 & 0.900 & 0.538  & 0.857 & 0.852 \\
TabNSA+Gemma & 0.687 & 0.913 & 0.571 & 0.824 & 0.825 \\

\midrule
\end{tabular}
}
\end{table*}




\section{Regression Task}
To evaluate model performance on a regression task, we compared six methods—MLP, XGBoost, CatBoost, TabNet, NSA only and our proposed TabNSA—using Root Mean Squared Error (RMSE) as the metric across three datasets: SA, TO, and CP. Lower RMSE indicates better performance. As shown in Table \ref{table:regression}, XGBoost achieved the best overall performance, with the lowest RMSE on the CP dataset and competitive scores across the others, resulting in the highest ranking. CatBoost closely followed, outperforming XGBoost on the SA dataset and ranking second overall. The MLP model struggled with relatively higher RMSE values, while TabNet exhibited significantly worse performance, particularly on the SA dataset. Our TabNSA model achieved the lowest RMSE on the TO dataset, demonstrating its effectiveness in certain cases. However, it reported a higher RMSE on the CP dataset, impacting its overall ranking. These results highlight the effectiveness of gradient-boosting models for tabular regression, while also showcasing TabNSA's strengths in specific domains.

\begin{table}[t]
    \centering
    \scriptsize 
    \setlength{\tabcolsep}{2pt} 
    \caption{Evaluation of Different Models for Regression task. Results shows RMSE}
    \label{table:regression}
    \begin{tabular}{lccc|c}
    \hline
    Methods  & SA & TO & CP & Rank $\downarrow$ \\
    \hline
    MLP & 0.607 & 0.037 & 7.895 & 5 \\
    XGBoost  & 0.185 & 0.0276 & \textbf{2.631} & \textbf{2} \\
    CatBoost  & 0.183 & 0.0281 & 2.848 & 2.33 \\
    TabNet & 3.2450 & 0.1050 & 3.405 & 5.33 \\ \midrule
    
    \textbf{NSA} & 0.3129 & \textbf{0.0273} & 14.2931 & 3.67 \\
     \textbf{TabNSA} & \textbf{0.1781} & 0.0291 & 2.9814 & \underline{2.67} \\
    
    \midrule
    \end{tabular}

\begin{tablenotes}[flushleft]
    \footnotesize 
    \item{Note: While TabNSA performs competitively, tree-based models remain superior for regression, highlighting opportunities for further model refinement.}
\end{tablenotes}
    
\end{table}

As shown in Table~\ref{table:regression}, TabNSA performs competitively but does not consistently surpass ensemble-based models like XGBoost and CatBoost on regression benchmarks. This gap may arise from architectural factors: TabNSA’s discrete attention and feature selection mechanisms are well-suited to classification tasks, where identifying sparse, instance-specific relevant features is critical. However, regression problems often involve smoother, continuous relationships between features and outputs, which may not benefit as directly from hard or coarse feature selection. Furthermore, the inductive bias of MLP-style backbones and sparse attention may underfit certain continuous-response distributions compared to tree-based ensembles that naturally partition the input space with high flexibility. Addressing this limitation is an important direction for future work. Possible improvements include incorporating soft attention over continuous-valued targets, hybridizing TabNSA with kernel-based output layers, or using continuous feature embeddings with uncertainty-aware attention mechanisms to better model real-valued outputs. Future work will explore adapting TabNSA for regression by incorporating soft-attention or hybrid continuous embedding mechanisms.

\section{Dataset}

We provide links to publicly available datasets used for benchmarking in this study. These datasets enable a comprehensive assessment of TabNSA's performance across various sectors and data types. Summary details are presented in Table~\ref{dataset}, with further descriptions provided below.

\noindent \textbf{Financial Decision-Making}:\\
Credit-g (CG) and Credit-Approval (CA) datasets, focused on credit risk evaluation and loan approval scenarios.

\noindent \textbf{Retail}:\\
Dresses-sale (DS) dataset, which includes detailed transaction records from dress sales.

\noindent \textbf{Demographic Analysis}:\\
Adult (AD) and 1995-Income (IO) datasets, covering income classification and census data.

\noindent \textbf{Specialized Industries}:\\
\noindent \textbf{Cylinder Bands (CB)} dataset, representing data from the manufacturing sector.  
\noindent \textbf{Blastchar (BL)} dataset, related to materials science.  
\noindent \textbf{Insurance Company (IC)} dataset, offering insights into the insurance domain.

\noindent \textbf{Blood} dataset contains data from a blood transfusion service in Taiwan, with 4 attributes describing 748 donors. The task is to predict whether a donor returns for another donation, with 178 positive cases.

\noindent \textbf{Bank} dataset includes 45,211 records from a direct marketing campaign by a Portuguese bank, with 16 features per record. The target is to predict term deposit subscriptions, with 5,289 positive labels.

\noindent \textbf{Credit-g} dataset describes 1,000 German credit applicants using 20 attributes. The binary classification task is to assess creditworthiness, identifying 700 individuals as having good credit.

\noindent \textbf{Heart} dataset aggregates data from four hospitals, comprising 918 patient records with 11 clinical features each. The goal is to detect coronary artery disease, with 508 positive cases.

\noindent \textbf{Income (Adult)} dataset consists of 48,842 records from the 1994 U.S. Census, each with 12 attributes. The task is to predict whether an individual's annual income exceeds \$50,000, with 11,687 positive examples.

\noindent \textbf{Jungle} dataset contains 44,819 end-game positions from Jungle Chess, used for specialized evaluation scenarios.

\begin{table}[htbp]
    \caption{Benchmark Dataset Links}
    \label{benchmark_datasets}
    \begin{tabularx}{\textwidth}{@{}lX@{}} 
        \toprule
        Dataset & URL \\
        \midrule
        Credit-G & \url{https://www.openml.org/search?type=data&status=active&id=31} \\
        Credit-Approval & \url{https://archive.ics.uci.edu/ml/datasets/credit+approval} \\
        Dress-Sales & \url{https://www.openml.org/search?type=data&status=active&id=23381} \\
        Adult & \url{https://www.openml.org/search?type=data&status=active&id=1590} \\ \midrule
        
        Cylinder-Bands & \url{https://www.openml.org/search?type=data&status=active&id=6332} \\
        Blastchar & \url{https://www.kaggle.com/datasets/blastchar/telco-customer-churn} \\
        Insurance-Co & \url{https://archive.ics.uci.edu/ml/datasets/Insurance+Company+Benchmark+(COIL+2000)} \\ 
        1995-Income & \url{https://www.kaggle.com/datasets/lodetomasi1995/income-classification} \\ \midrule
       
        Bank & \url{https://www.openml.org/search?type=data&status=active&id=45065} \\
        Blood & \url{https://www.openml.org/search?type=data&status=active&id=1464} \\
        Diabetes & \url{https://www.kaggle.com/datasets/uciml/pima-indians-diabetes-database} \\ 
        Heart & \url{https://www.kaggle.com/datasets/fedesoriano/heart-failure-prediction} \\
        Jungle & \url{https://www.openml.org/search?type=data&status=active&id=41027} \\ \midrule

        ImageSegmentation & \url{https://www.openml.org/search?type=data&status=active&id=36} \\
        ForestCovertype  & \url{https://www.openml.org/search?type=data&status=active&id=150} \\
        
        \bottomrule
    \end{tabularx}
\end{table}

\section{Sensivity Analysis}

The provided plot of AUC versus the number of Optuna trials illustrates the sensitivity of TabNSA's performance to the hyperparameter search budget. The curve shows a sharp increase in performance within the initial 20-40 trials, after which the rate of improvement diminishes significantly. By approximately 60-100 trials, the performance has fully plateaued, indicating that further searching yields negligible gains.

This trajectory demonstrates two key findings. First, Optuna’s Bayesian sampler efficiently converges towards optimal hyperparameter configurations, rapidly capturing the model's inductive biases. Second, and more practically, this analysis reveals that a modest budget of approximately 50 trials is sufficient to achieve over 95\% of the model's peak performance. This provides a clear and practical guideline for researchers and practitioners seeking to apply TabNSA effectively without incurring prohibitive computational costs.

\begin{figure*}[!t]
\centering
\caption{\textbf{Search Budget Sensitivity Curve:} The figure shows the best validation AUC achieved by TabNSA as a function of the number of Optuna trials, across three datasets (DS, CG, and CA). The curve illustrates how model performance improves with increased hyperparameter search budget.}
\resizebox{0.8\textwidth}{!}{
    \includegraphics{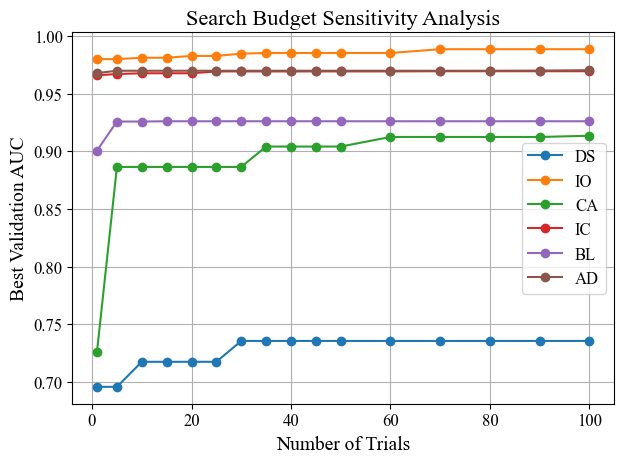}
}
\label{fig:sensivity}
\end{figure*}

\section{Component-wise ablation of sparse attention parameters}
\begin{figure*}[!t]
  \centering
  \begin{subfigure}[b]{0.48\textwidth}
    \includegraphics[width=\textwidth]{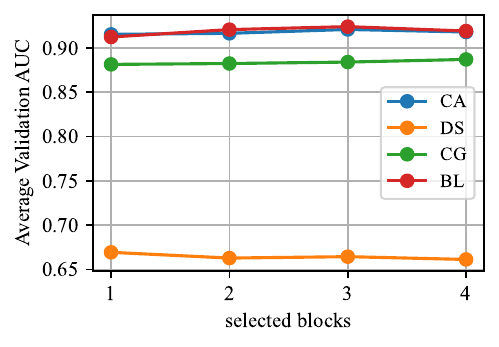}
    \caption{Impact of selected blocks on Validation Performance}
    \label{fig:selblock}
  \end{subfigure}
  \hfill
  \begin{subfigure}[b]{0.48\textwidth}
    \includegraphics[width=\textwidth]{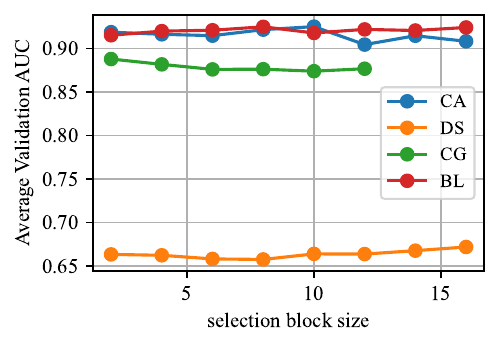}
    \caption{Impact of selection block size on Validation Performance}
    \label{fig:selsize}
  \end{subfigure}

  \vspace{1ex}
  \begin{subfigure}[b]{0.48\textwidth}
    \includegraphics[width=\textwidth]{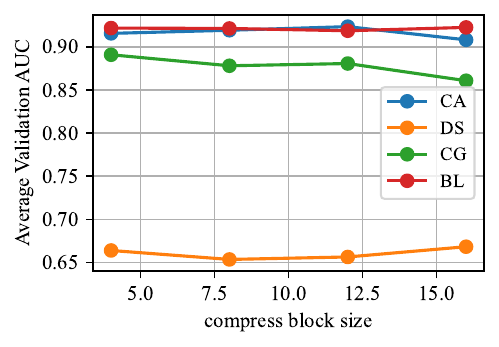}
    \caption{Impact of compress block size on Validation Performance}
    \label{fig:compsize}
  \end{subfigure}
  \hfill
  \begin{subfigure}[b]{0.48\textwidth}
    \includegraphics[width=\textwidth]{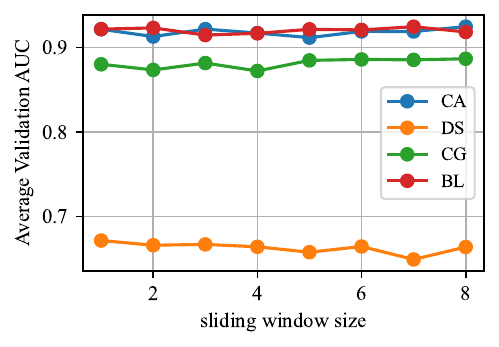}
    \caption{Impact of sliding window size on Validation Performance}
    \label{fig:slidewin}
  \end{subfigure}

  \caption{Component-wise ablation of sparse attention parameters. Each plot shows the effect of varying one parameter : \textit{number\_of\_selected blocks} (a), \textit{selection\_block\_size} (b),  \textit{compress\_block\_size} (c) and \textit{sliding\_window\_size} (d) on the average validation AUC across four datasets. The results highlight the importance of careful tuning and indicate the optimal value ranges for each parameter.}
  \label{fig:all_steps}
\end{figure*}

We conducted a component-wise ablation study to analyze each sparse attention parameter’s contribution to model performance. The results can be seen in Fig. \ref{fig:all_steps}. Specifically, we varied one parameter at a time while keeping others fixed and evaluated the results on four validation sets. For \textit{compress block size}, a value of 10 achieved the highest average validation AUC ($\sim$0.87), with performance declining at smaller or larger values. Selecting 3 blocks for the \textit{number of selected blocks} yielded optimal performance ($\sim$0.86), while fewer or more blocks reduced effectiveness. A \textit{selection block size} of 10 was optimal ($\sim$0.86), with deviations in either direction slightly lowering performance. Lastly, a \textit{sliding window size} of 6 delivered the best performance ($\sim$0.88), demonstrating the importance of capturing appropriate local context. With this information, we can constrain the search space in the first-stage optimization to accelerate convergence toward the optimal AUC. However, these hyperparameters should still be searched jointly to effectively optimize the model for each dataset.

\end{appendices}
\end{document}